\documentclass[10pt,twocolumn,letterpaper]{article}

\usepackage{cvpr}              

\usepackage[T1]{fontenc}
\usepackage[utf8]{inputenc}
\usepackage{lmodern}

\definecolor{cvprblue}{rgb}{0.21,0.49,0.74}
\usepackage[pagebackref,breaklinks,colorlinks,allcolors=cvprblue]{hyperref}
\usepackage{graphicx}   
\usepackage{booktabs}   
\usepackage{multirow}
\usepackage{makecell}   
\usepackage{amssymb}    
\usepackage{pifont}     
\usepackage{import}
\usepackage{diagbox}
\usepackage{fontawesome}
\usepackage{siunitx}
\usepackage{arydshln}
\usepackage{makecell}
\usepackage{comment}
\usepackage{tabularx}

\newcommand{\modRGB}{\faCamera}
\newcommand{\modDepth}{\faCube}
\newcommand{\modForce}{\faHandRockO}
\newcommand{\modHaptic}{\faHandPaperO}
\newcommand{\modHand}{\faHandPeaceO}
\newcommand{\modAudio}{\faVolumeUp}
\newcommand{\modJoints}{\faCogs}
\newcommand{\modEye}{\faEye}
\newcommand{\modthreeD}{\faCubes}
\newcommand{\modLang}{\faCommentO}

\newcommand{\PAR}[1]{\vskip 0pt \noindent{\bf #1~}}
\def\paperID{} 
\def\confName{CVPR}
\def\confYear{2026}

\makeatletter
\renewenvironment{abstract}{
    \centerline{\large\bfseries Abstract}
    \vspace{-15pt}
    \itshape
}{}
\makeatother

\title{Hoi! - A Multimodal Dataset for Force-Grounded, Cross-View Articulated Manipulation}

\author{
Tim Engelbracht$^{1,\dagger}$ \quad
René Zurbrügg$^{1}$ \quad
Matteo Wohlrapp$^{2}$ \quad
Martin Büchner$^{3}$ \quad \\
Abhinav Valada$^{3}$ \quad
Marc Pollefeys$^{1,4}$ \quad
Hermann Blum$^{5}$ \quad
Zuria Bauer$^{1}$ \\
\\
$^{1}$ETH Zurich \quad
$^{2}$Technical University of Munich \quad
$^{3}$University of Freiburg \\
$^{4}$Microsoft \quad
$^{5}$University of Bonn \\
\\}

\begin{document}

\twocolumn[{%
\renewcommand\twocolumn[1][]{#1}%
\maketitle 

     
\begin{center}
    \vspace*{-32pt}\includegraphics[width=.8\textwidth]{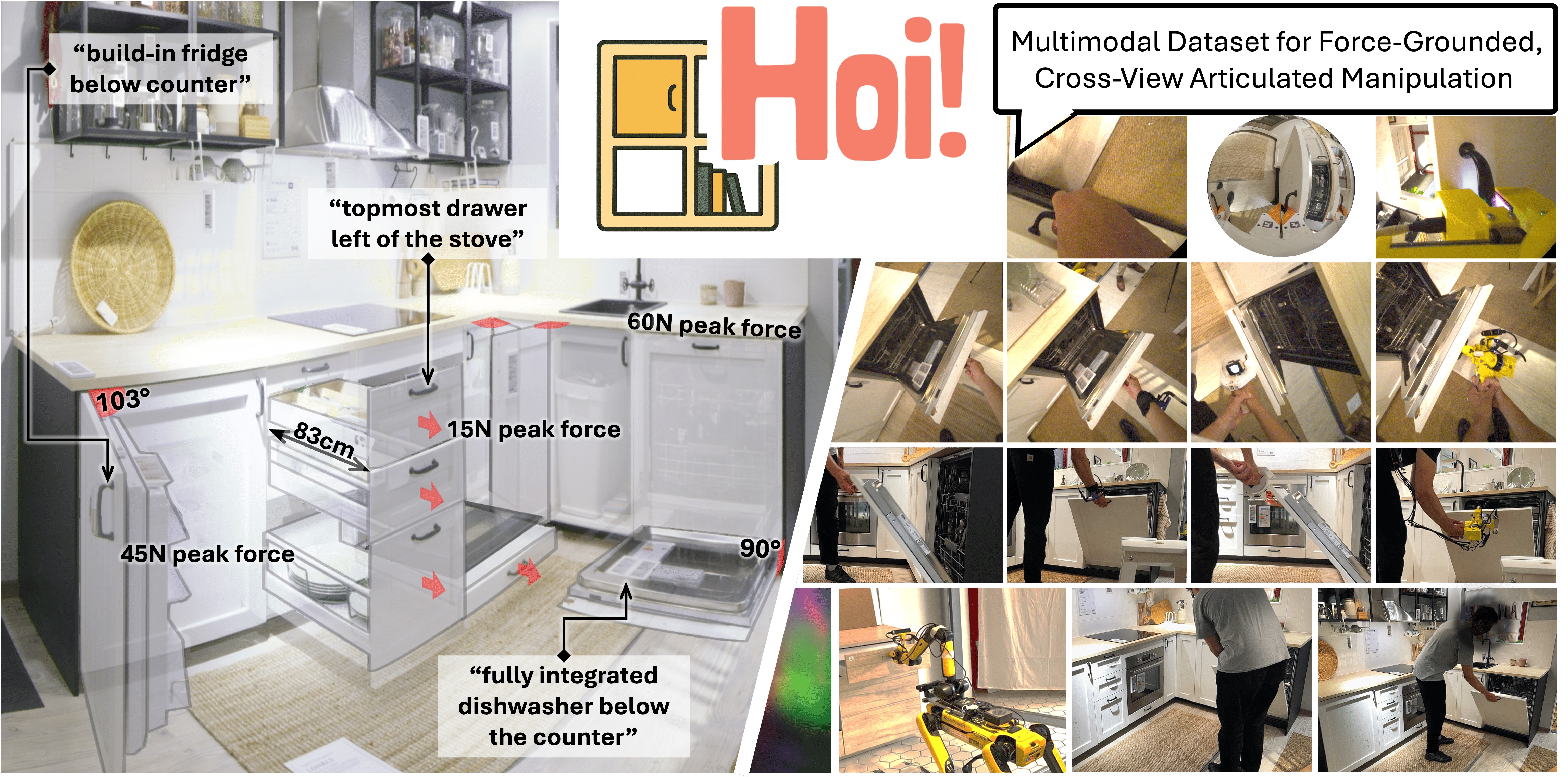}
  \captionof{figure}{
    \small
    \textbf{Overview of the Hoi! Dataset}: 
    A multimodal dataset for force-grounded, cross-view articulated manipulation in wild indoor environments. The dataset captures human interactions with common articulated objects (drawers, doors, fridges, dishwashers) with synchronized RGB, depth, force, tactile sensing, and multi-view videos from egocentric and exocentric perspectives. Each interaction is annotated with articulation parameters (e.g. axis, type), supporting research on multimodal perception, manipulation learning, and embodied reasoning. 
  }\label{fig:teaser}
\end{center}

}]

\begin{abstract} 
{\fontsize{9.9pt}{11.2pt}\selectfont
We present a dataset for force-grounded, cross-view articulated manipulation that couples what is seen with what is done and what is felt during real human interaction. 
The dataset contains \textbf{3048} sequences across \textbf{381} articulated objects in \textbf{38} environments. 
Each object is operated in four embodiments - (i) human hand, (ii) human hand with a wrist-mounted camera, (iii) handheld UMI gripper, and (iv) a custom Hoi! gripper, where the tool embodiment provides end-effector forces and tactile sensing.
Our dataset offers a holistic view of interaction understanding from video, enabling researchers to evaluate how well methods transfer between human and robotic viewpoints, but also investigate underexplored modalities such as interaction forces. The Project Website can be found \href{https://timengelbracht.github.io/Hoi-Dataset-Website/}{here}.\newline
{$^{\dagger}$\tt\small Corresponding author (tengelbracht@ethz.ch)}
\vspace{-15pt}}
\end{abstract}

    
\section{Introduction}
\label{sec:intro}
In recent years, computer vision has moved from purely perceptive domains to a dynamic and interactive era, with research increasingly aiming to interpret how objects can be used or interacted with.
%
%
Progress in this direction has been largely fueled by data-driven methods and the availability of large-scale datasets such as \cite{EpicKitchen_Damen2022, HDEpic_perrett2025hd, grauman2022ego4d,grauman2024egoexo4dunderstandingskilledhuman} capturing diverse human and object interactions.
Such datasets enable learning-based models to generalize across different tasks and environments, thereby moving toward generalizable, zero-shot actionable understanding.
%
A closer examination, however, reveals a fundamental discrepancy in the nature of interactions studied across different research domains. While human–centric video datasets emphasize long-horizon activities such as cooking, furniture assembly, and sports, robotics datasets predominantly target short-horizon primitives like pick-and-place, wiping, or drawer opening. This data gap makes it hard to investigate interesting transfer questions: Do interaction-force predictors generalize to human videos? Do articulation-tracking methods remain effective from an exocentric robotic viewpoint? Can interactions demonstrated by a human hand be re-targeted to a two-finger robotic gripper?
Among the many forms of interaction, articulated furniture provides an especially rich yet understudied case. Despite its prevalence in everyday human activity, curated video data of people interacting with furniture is sparse. 
On the robotics side, articulated furniture manipulation is emerging as a tractable challenge, with perception identified as the main obstacle to further progress~\cite{gupta2025openingarticulatedstructuresreal}. Existing articulation datasets~\cite{halacheva2024articulate3d, delitzas2024scenefun3d} offer valuable labels but are largely constructed from static scans, lacking the paired motion data needed to ground these annotations in real interactions.
To bridge this gap, we introduce a dataset for force-grounded, cross-view, multi-embodiment manipulation of articulated furniture. 
Each object is operated in four embodiments: (i) a human hand, (ii) a human hand with a wrist-mounted camera, (iii) a handheld UMI gripper, and (iv) a custom Hoi! gripper equipped with tactile sensors and a force-torque sensor.
To anchor perception and interaction to scene geometry, we scan each environment before and after manipulation with a laser scanner, yielding dense 3D point clouds that capture both static structure and interaction-induced geometry changes. We augment the raw data with interaction annotations, including sequence-level state-change labels (e.g., open/close progression), object metadata (category, articulation type), and joint parameters. This combination of forces, egocentric cues, third-person context, and scene-level ground truth provides a unified basis for studying the multimodality and entire sensor range of interaction on everyday articulated objects.
In summary, our contributions are:
\begin{itemize}

\item A data capture pipeline for human demonstrations that couples multi-view RGB-D observations with end-effector forces and tactile signals, enabled by a novel hand-held robotic gripper for recording in-the-wild interaction forces.

\item A multi-view dataset for articulated manipulation: \textbf{3048} sequences over \textbf{381} objects and \textbf{38} environments, spanning four main embodiments.  Each sequence provides time-aligned vision, pose, force, haptic streams and precise scene-level ground truth.

\item Interaction-centric annotations supporting benchmarks in force-from-vision, articulation estimation, cross-view transfer, and state-change prediction.

\end{itemize}

\newcommand{\cmark}{\ding{51}}%
\newcommand{\xmark}{\ding{55}}%

\begin{table*}[!h]
\vspace{-10pt}
\caption{\textbf{Commonly used datasets for human interactions and articulated environments.}
\textit{Views:} Numbers indicate available camera viewpoints; columns show \textbf{Egocentric (Ego)}, \textbf{Exocentric (Exo)}, and \textbf{Wrist-mounted (Wrist)}.
\textit{Embodiments:} Columns show \textbf{Human (H)}, \textbf{Robot (R)}, and \textbf{Tool/Gripper (T)}.
\textit{Modalities:}
\modRGB~RGB, 
\modDepth~Depth, 
\modForce~Force/Torque, 
\modHaptic~Haptic/Tactile, 
\modHand~Hand Tracking, 
\modAudio~Audio, 
\modJoints~Joint States, 
\modEye~Eyetracking, 
\modthreeD~3D Model / Digital Twin, 
\modLang~Language. \label{tab:overview_datasets}}

\centering
\vspace*{-8pt}
\resizebox{\linewidth}{!}{%
\begin{tabular}{l l l l ccc ccc l}
\toprule
\textbf{Dataset} & \textbf{Class} & \textbf{Tasks/Objects} & \textbf{Environment} &
\multicolumn{3}{c}{\textbf{Views}} & \multicolumn{3}{c}{\textbf{Embodiments}} & \textbf{Modalities} \\
\cmidrule(lr){5-7} \cmidrule(lr){8-10}
& & & & \textbf{Ego} & \textbf{Exo} & \textbf{Wrist} & \textbf{H} & \textbf{R} & \textbf{T} & \\
\midrule
RBO~\cite{RBO_dataset} & articulation & \parbox{4cm}{$\sim$\SI{1}{h} of interactions,\\ 14 articulated objects} & ~ &
\xmark & {1×\scriptsize}\cmark & \xmark &
\cmark & \xmark & \xmark &
\modRGB $\textrm{\modthreeD}_\textrm{articulated meshes}$ \\

RH20T~\cite{fang2023rh20tcomprehensiveroboticdataset} & \parbox{4cm}{pick \& place, toy playing \\ very few articulations} & $\sim$\SI{916}{h}, 110k demos & tabletop &
\xmark & {6×\scriptsize}\cmark & \cmark &
\cmark & {4×\scriptsize}\cmark & \xmark &
\modRGB \modDepth \modForce \modAudio \modJoints \\

EgoExo4D~\cite{grauman2024egoexo4dunderstandingskilledhuman} & cooking, assembly, sports & \SI{1286}{h}, 690 actions & 123 scenes &
\cmark & {4×\scriptsize}\cmark & \xmark &
\cmark & \xmark & \xmark &
\modRGB \modDepth \modForce \\

ForceMimic~\cite{liu2025forcemimicforcecentricimitationlearning} & cooking & \SI{30}{k} zucchini peeling sequences & tabletop &
\xmark & {1×\scriptsize}\cmark & \xmark &
\cmark & \cmark & \cmark &
\modRGB \modDepth \modForce \\

ArticuBot~\cite{wang2025articubotlearninguniversalarticulated} & articulation & 322 articulated parts, 42k demos & simulation &
 & {Multi.\scriptsize} &  &
\xmark & \cmark & \xmark &
\modRGB \modDepth $\textrm{\modthreeD}_\textrm{sim assets}$ \modJoints  \\

Kaiwu~\cite{Kaiwu_ren2025} & assembly & \SI{40}{h}, 30 objects and \SI{11}{k} interactions& tabletop &
\xmark & {1×\scriptsize}\cmark & \xmark &
\cmark & \xmark & \xmark &
\modRGB \modDepth \modHand $\textrm{\modHaptic}_\textrm{haptic glove}$ \modAudio \modEye \\

EpicKitchens~\cite{EpicKitchen_Damen2022} & cooking & \SI{100}{h} of cooking activities & 48 kitchens &
\cmark & \xmark & \xmark &
\cmark & \xmark & \xmark &
\modRGB \modAudio \modLang \\

DROID~\cite{Droid_khazatsky2024} & \parbox{4cm}{articulation, pick \& place, \\cleaning} & \SI{188}{h} teleoperation & 52 buildings &
\xmark & {2×\scriptsize}\cmark & \cmark &
\xmark & {3×\scriptsize}\cmark & \xmark &
\modRGB \modDepth \modJoints \\

Arti4D~\cite{ARTI4D_werby2025} & articulation & \SI{1}{h} of interacting with 85 objects & 4 scenes &
\cmark & \xmark & \xmark &
\cmark & \xmark & \xmark &
\modRGB \modDepth $\textrm{\modthreeD}_\textrm{reconstruction}$ \\

AgiBot World~\cite{AGIBOT_bu2025agibot} & pick \& place & $\sim$\SI{16}{kh} teleoperation & 106 scenes &
\cmark & \xmark & \cmark &
\xmark & \cmark & \xmark &
\modRGB \modDepth $\textrm{\modHaptic}_\textrm{visuo-tactile}$ \modJoints \\

OpenFunGraph~\cite{OpenFunGraphzhang2025open} & articulation & \SI{.5}{h} of interaction with 201 elements & 14 scenes &
\cmark & \xmark & \xmark &
\cmark & \xmark & \xmark & \modRGB
$\textrm{\modthreeD}_\textrm{3D scans}$ \\

HDEpic~\cite{HDEpic_perrett2025hd} & cooking & \SI{41}{h} of cooking activities & 9 kitchens &
\cmark & \xmark & \xmark &
\cmark & \xmark & \xmark &
\modRGB \modAudio \modEye \modHand $\textrm{\modthreeD}_\textrm{digital twin}$ \modLang \\
\hdashline

Hoi! (ours) & articulation & \parbox{4cm}{\SI{48}{h} of interactions,\\ 381 articulated parts} & 38 scenes &
\cmark & {2×\scriptsize}\cmark & \cmark &
\cmark & (\cmark) & {2×\scriptsize}\cmark &
\makecell{\modRGB \modDepth \modForce $\textrm{\modHaptic}_\textrm{digit tactile}$ \modHand \\ $\textrm{\modJoints}_\textrm{gripper only}$ $\textrm{\modthreeD}_\textrm{3D scans}$} \\

\bottomrule
\end{tabular}}
\smallskip
\vspace*{-8pt}
\end{table*}

\begin{figure*}[!t]
  \centering
    \newcommand{\sz}{0.34}
    \newcommand{\hw}{4.5cm}
    \resizebox{1.0\linewidth}{!}{%
    \begin{tabular}{cccccccc}
        \\  \vspace{-1pt}
        \includegraphics[width=0.33\linewidth]{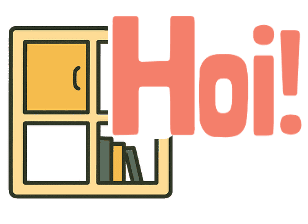} &
        \includegraphics[width=\sz\linewidth, height=\hw]{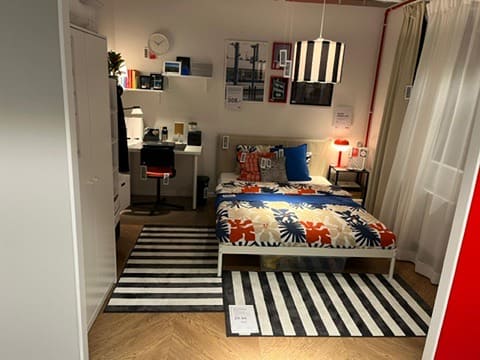} &
        \includegraphics[width=\sz\linewidth, height=\hw]{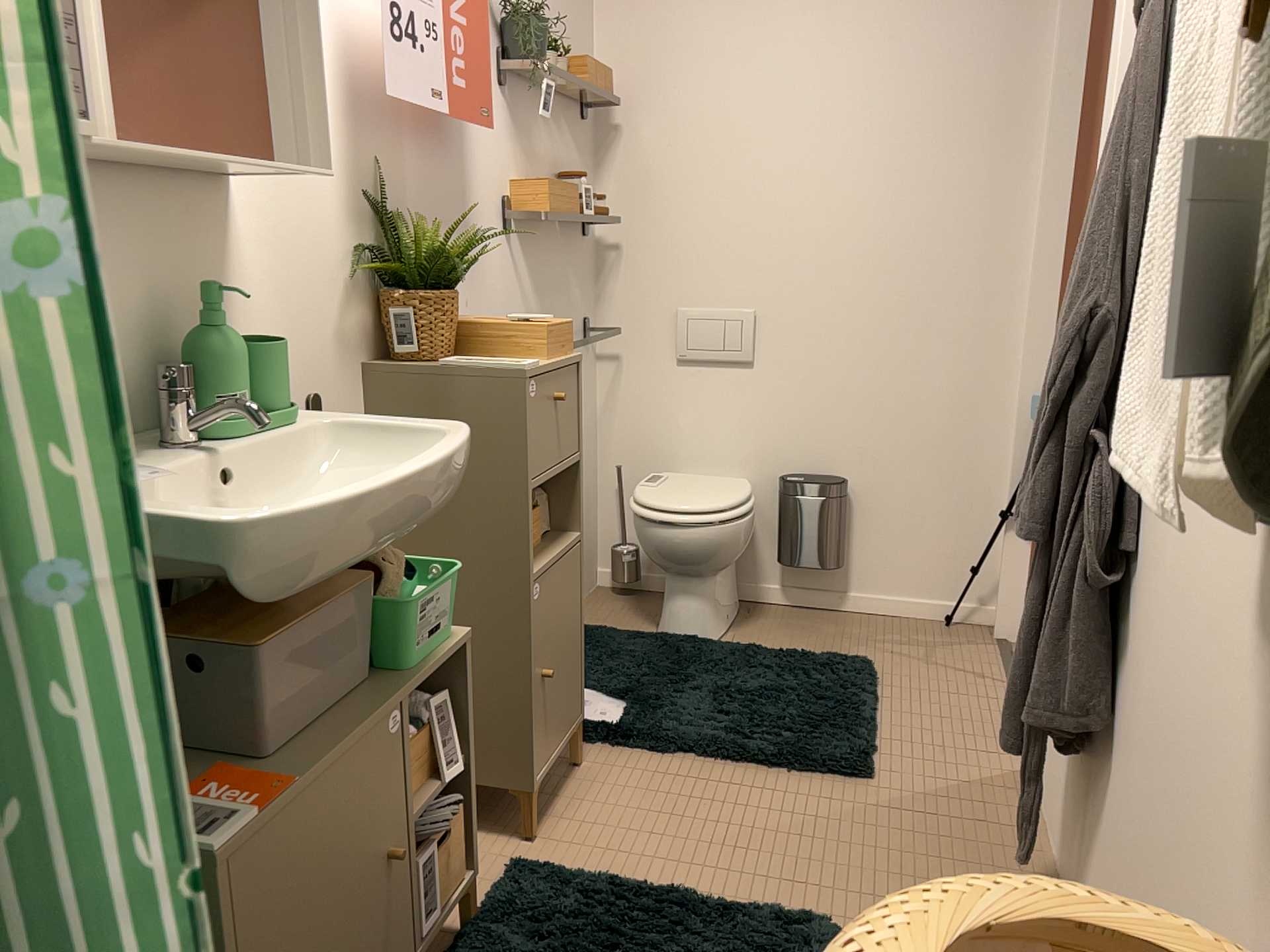} &
        \includegraphics[width=\sz\linewidth, height=\hw]{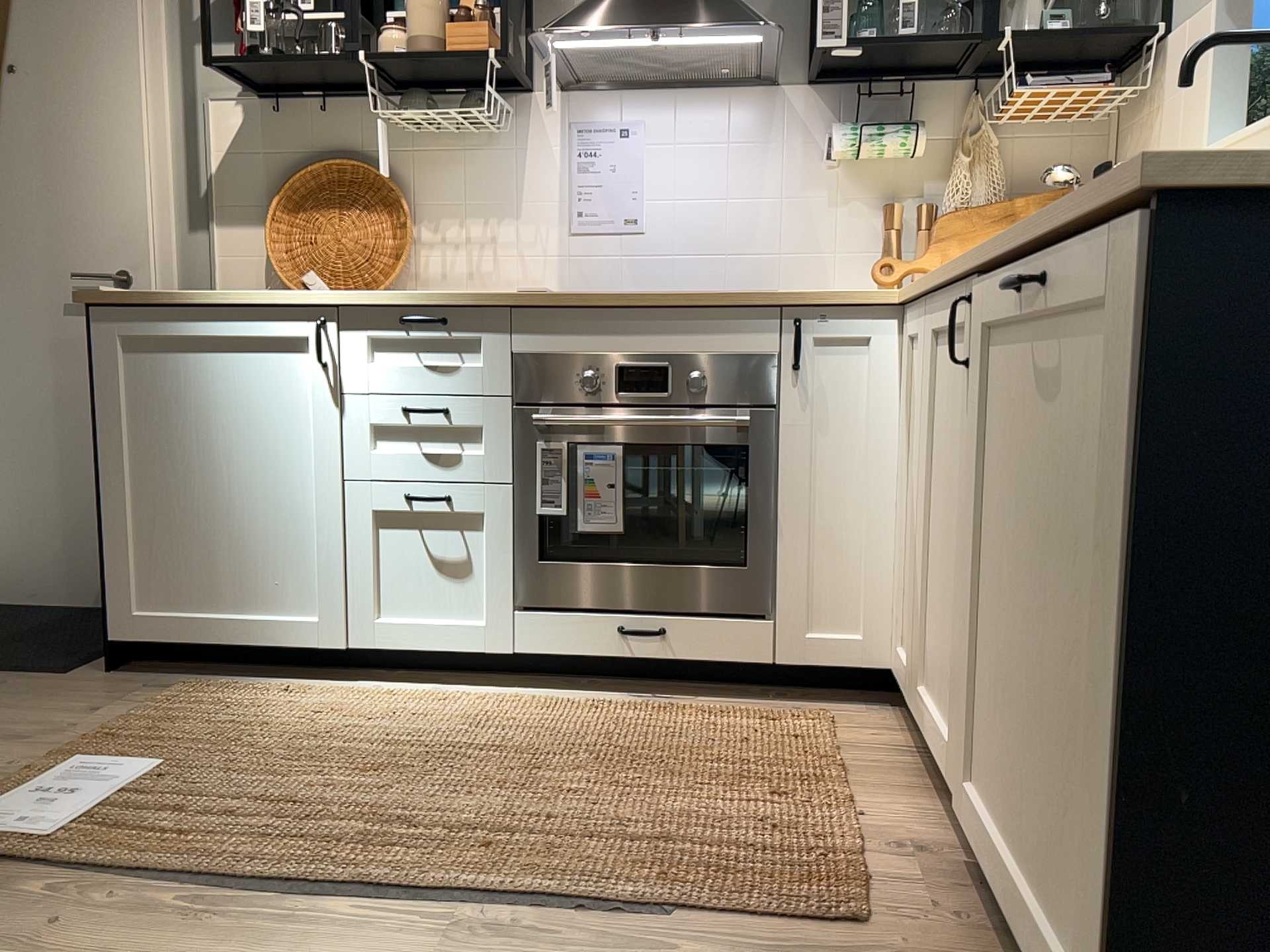} &
        \includegraphics[width=\sz\linewidth, height=\hw]{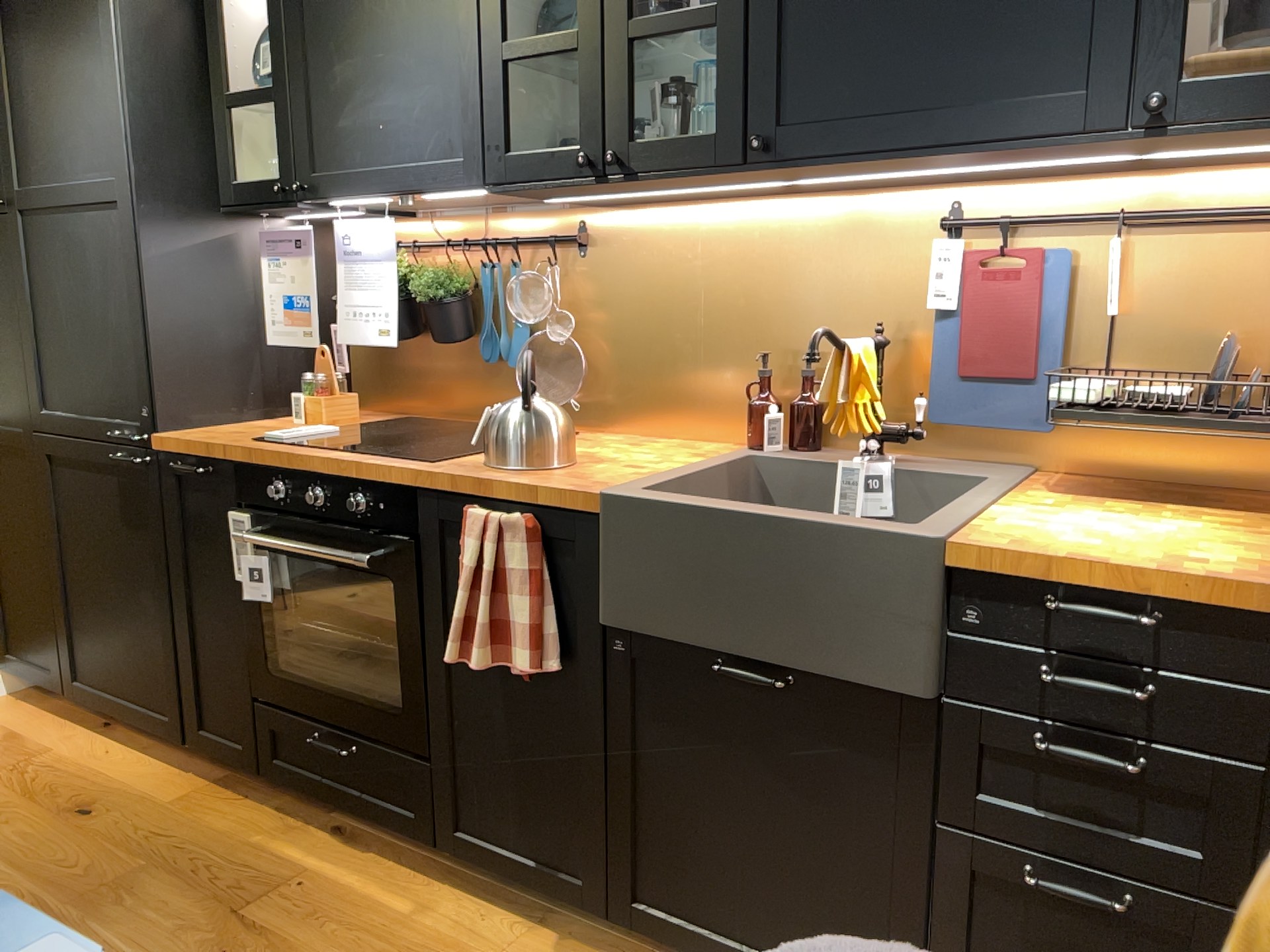} &
        \includegraphics[width=\sz\linewidth, height=\hw]{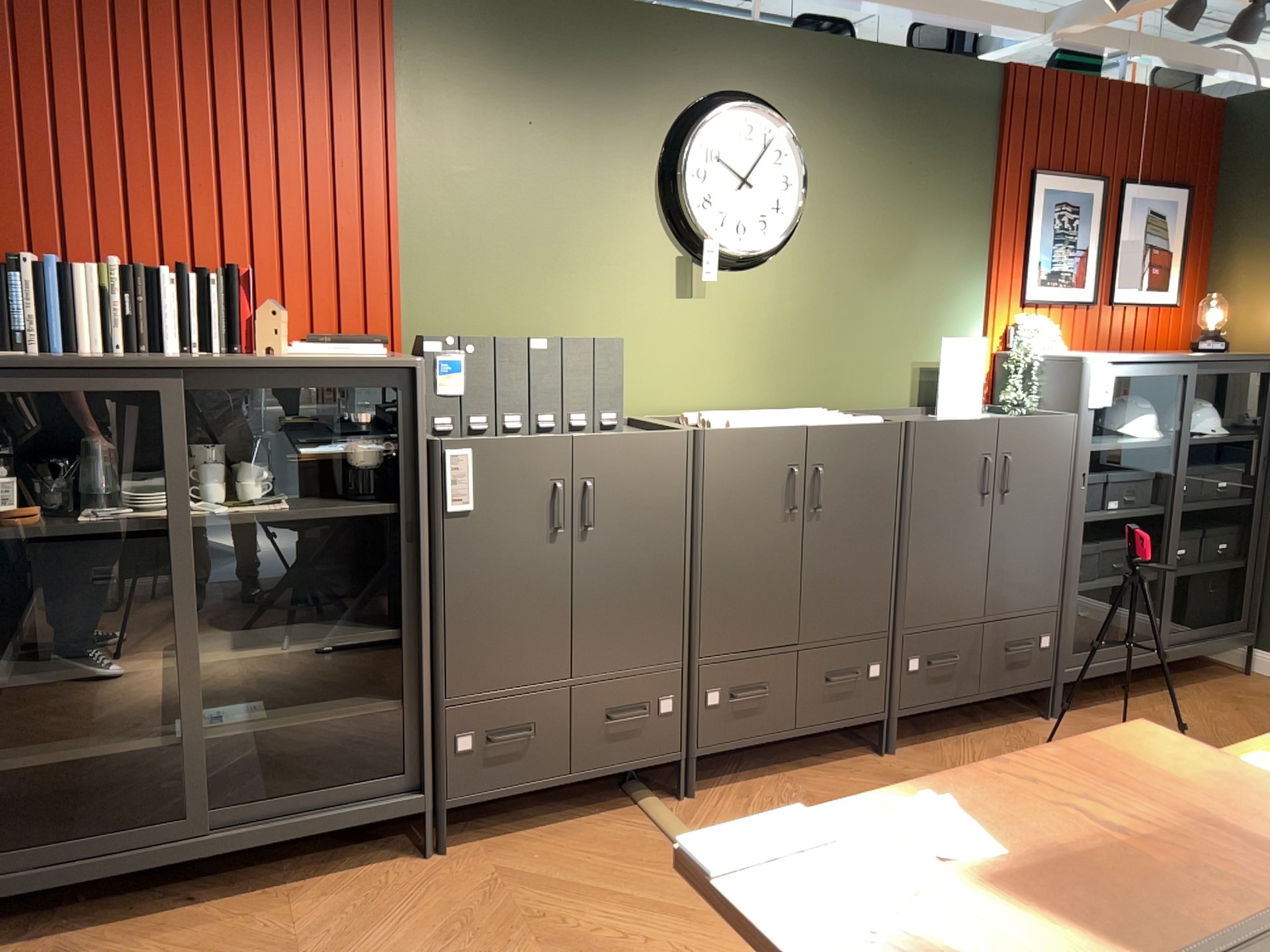} &
        \includegraphics[width=\sz\linewidth, height=\hw]{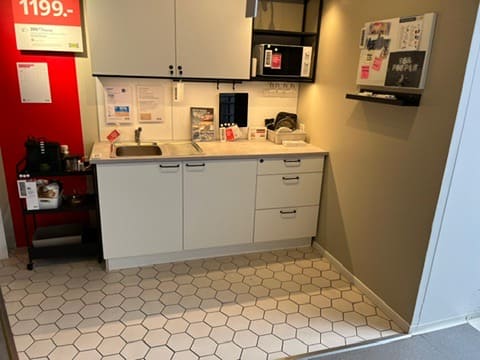} &
        \includegraphics[width=\sz\linewidth, height=\hw]{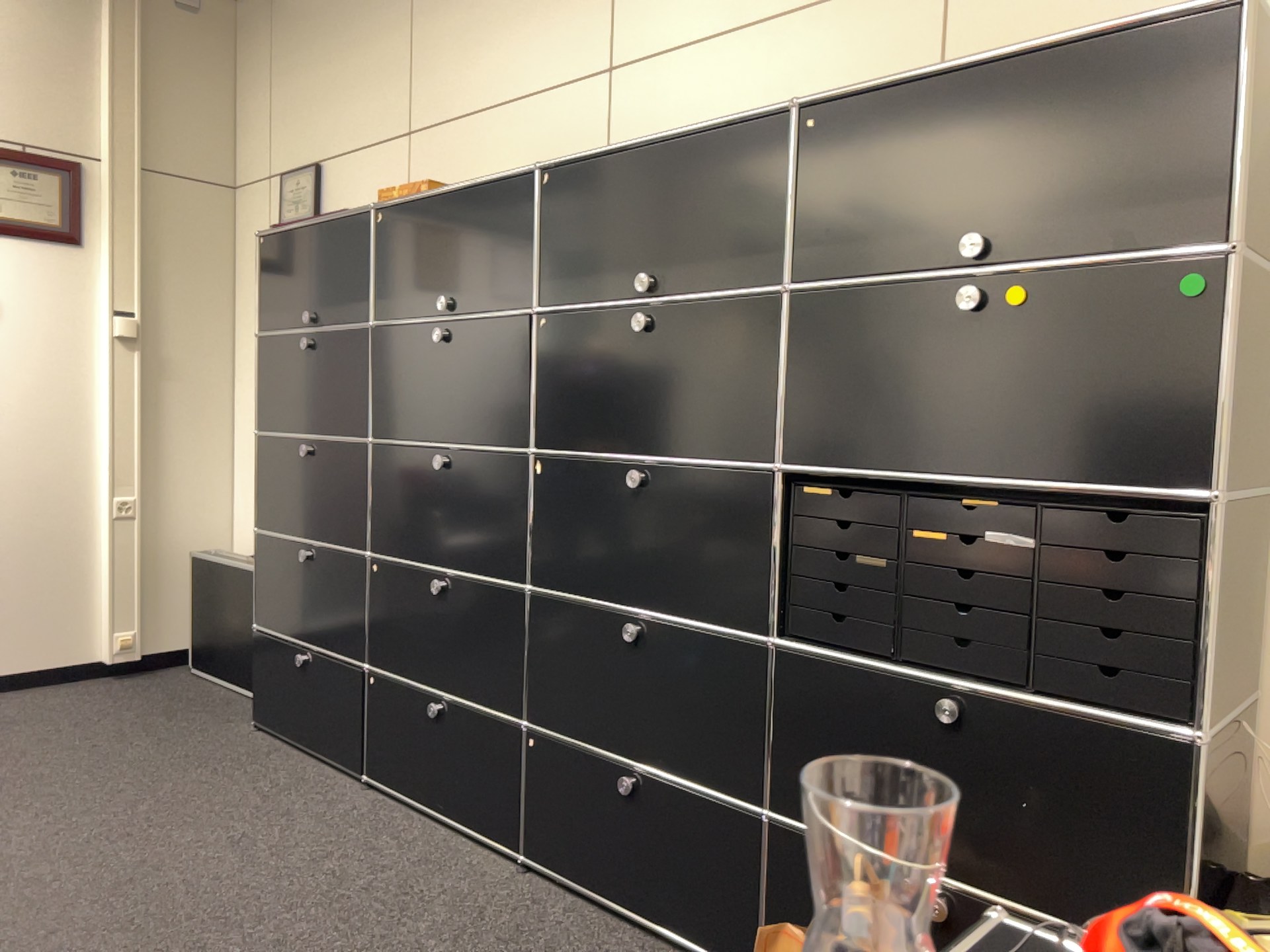} 
        \\ 
        \includegraphics[width=\sz\linewidth, height=\hw]{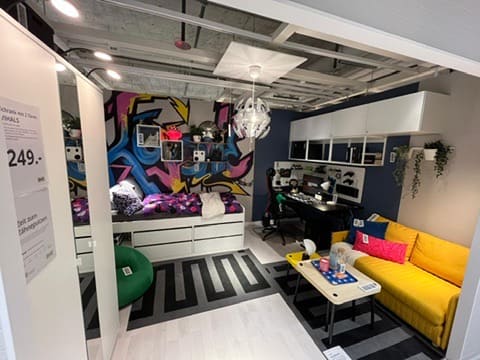} &
        \includegraphics[width=\sz\linewidth, height=\hw]{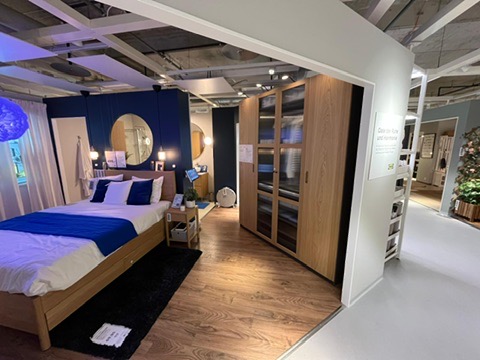} &
        \includegraphics[width=\sz\linewidth, height=\hw]{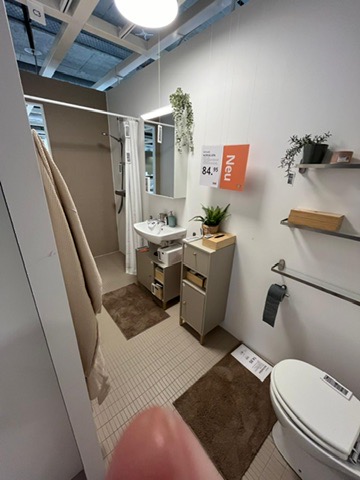} &
        \includegraphics[width=\sz\linewidth, height=\hw]{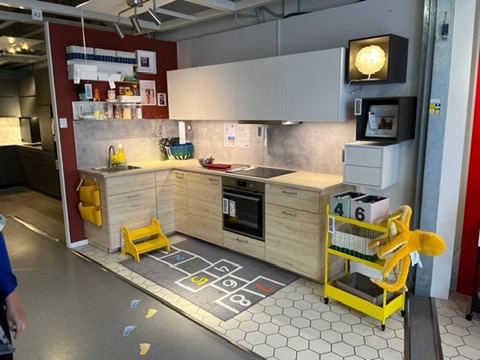} &
        \includegraphics[width=\sz\linewidth, height=\hw]{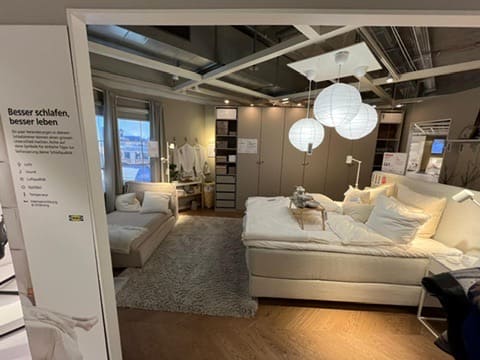} &
        \includegraphics[width=\sz\linewidth, height=\hw]{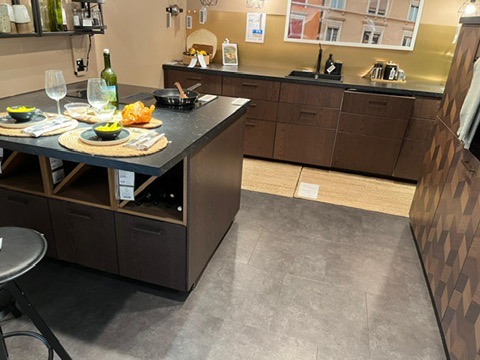} &
        \includegraphics[width=\sz\linewidth, height=\hw]{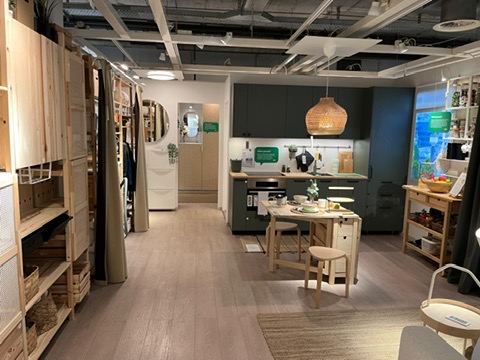} &
        \includegraphics[width=\sz\linewidth, height=\hw]{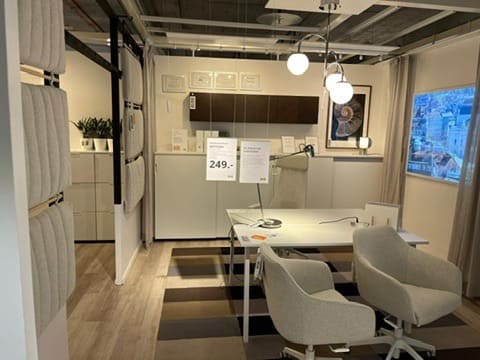} 
         \\
        \includegraphics[width=\sz\linewidth, height=\hw]{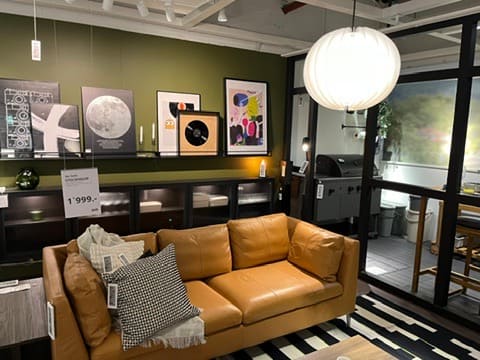} &
        \includegraphics[width=\sz\linewidth, height=\hw]{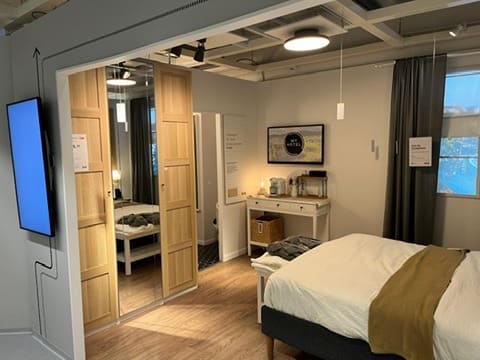} &
        \includegraphics[width=\sz\linewidth, height=\hw]{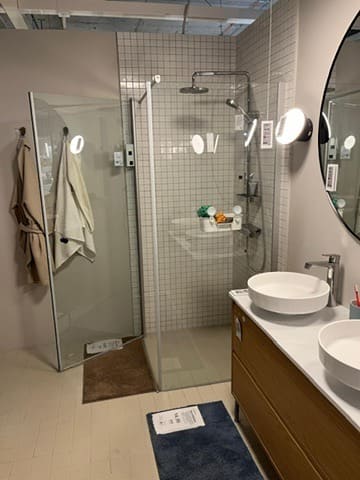} &
        \includegraphics[width=\sz\linewidth, height=\hw]{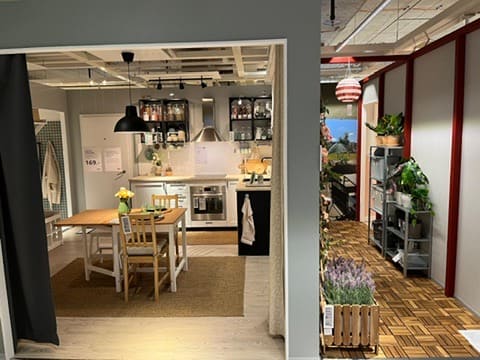} &
        \includegraphics[width=\sz\linewidth, height=\hw]{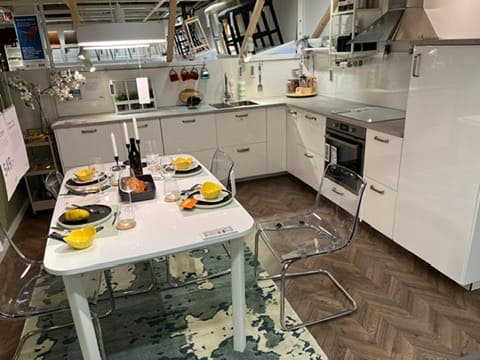} &
        \includegraphics[width=\sz\linewidth, height=\hw]{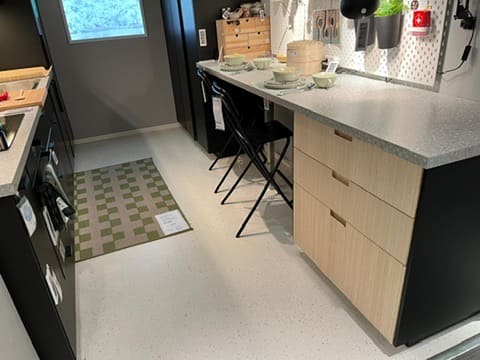} &
        \includegraphics[width=\sz\linewidth, height=\hw]{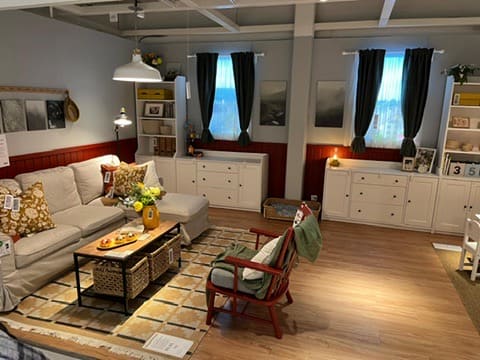} &
        \includegraphics[width=\sz\linewidth, height=\hw]{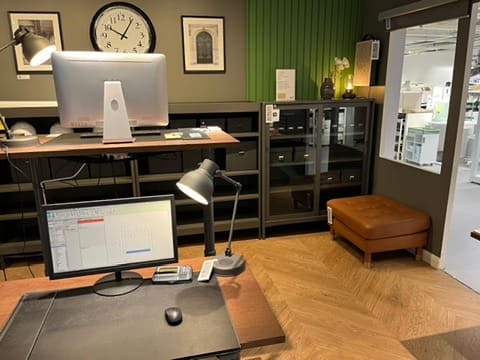} 
        \\ 
        \includegraphics[width=\sz\linewidth, height=\hw]{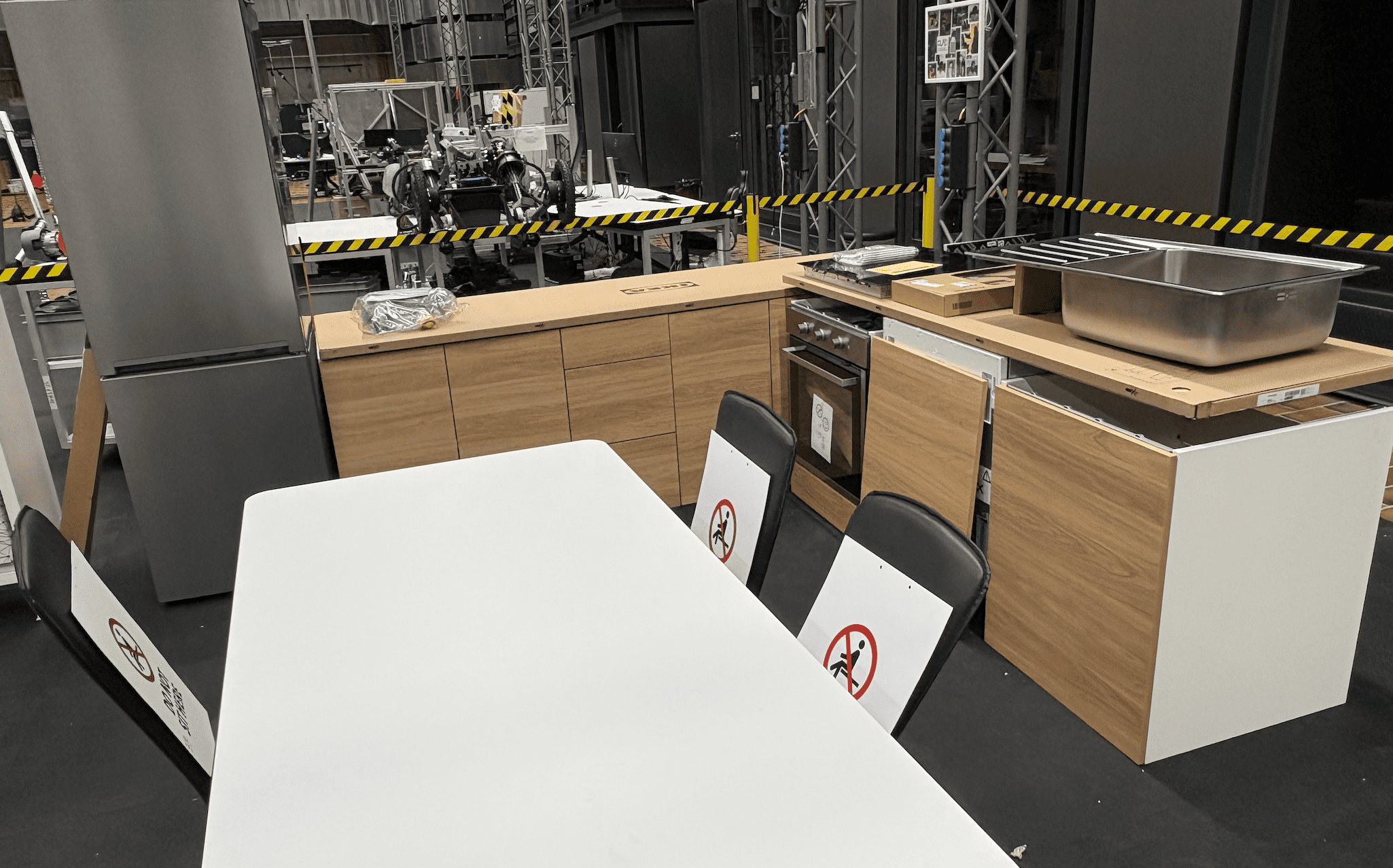} &
        \includegraphics[width=\sz\linewidth, height=\hw]{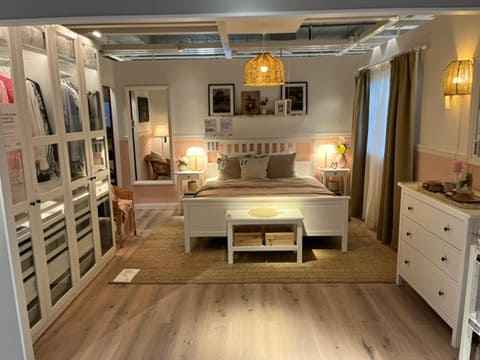} &
        \includegraphics[width=\sz\linewidth, height=\hw]{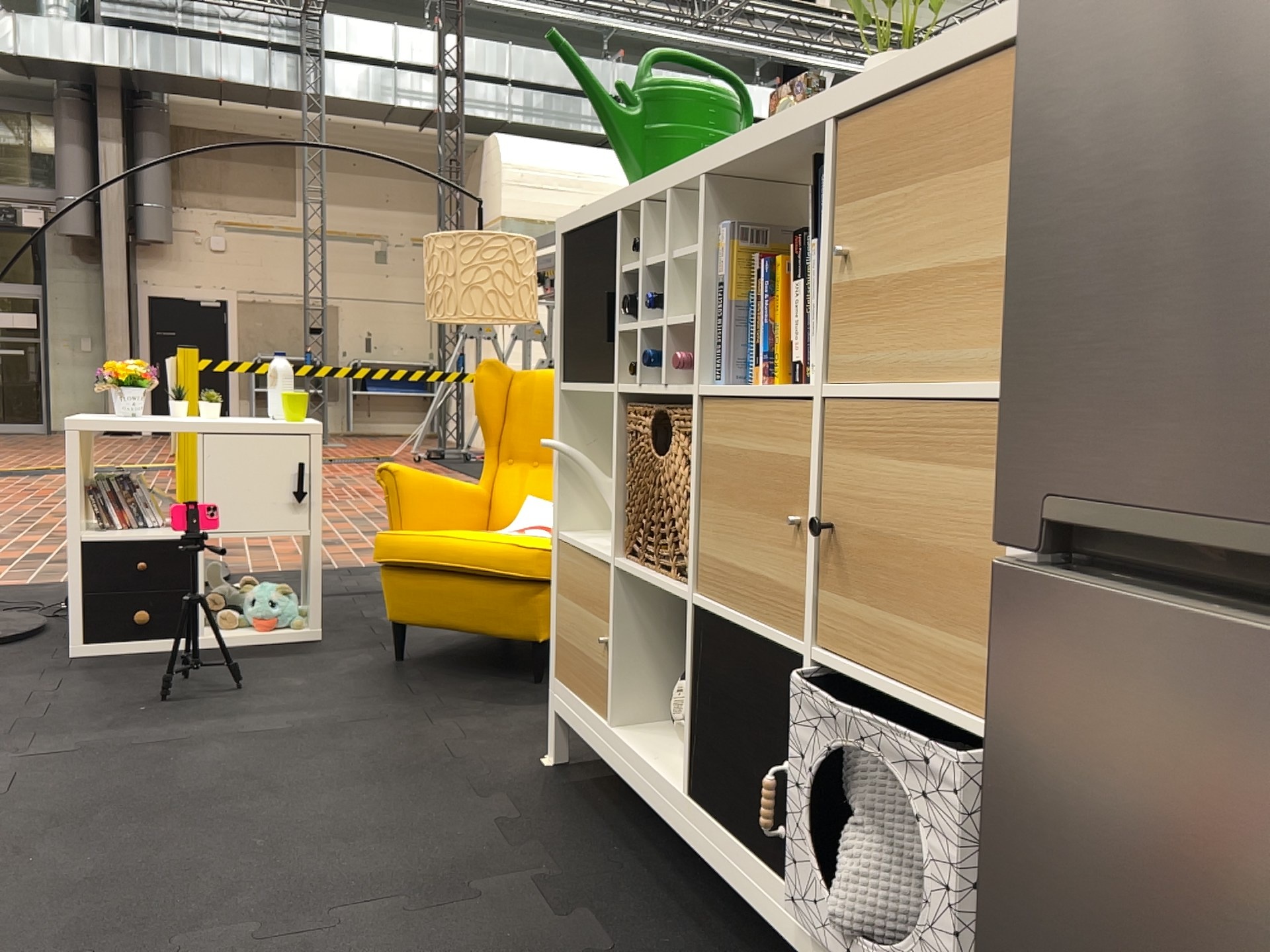} &
        \includegraphics[width=\sz\linewidth, height=\hw]{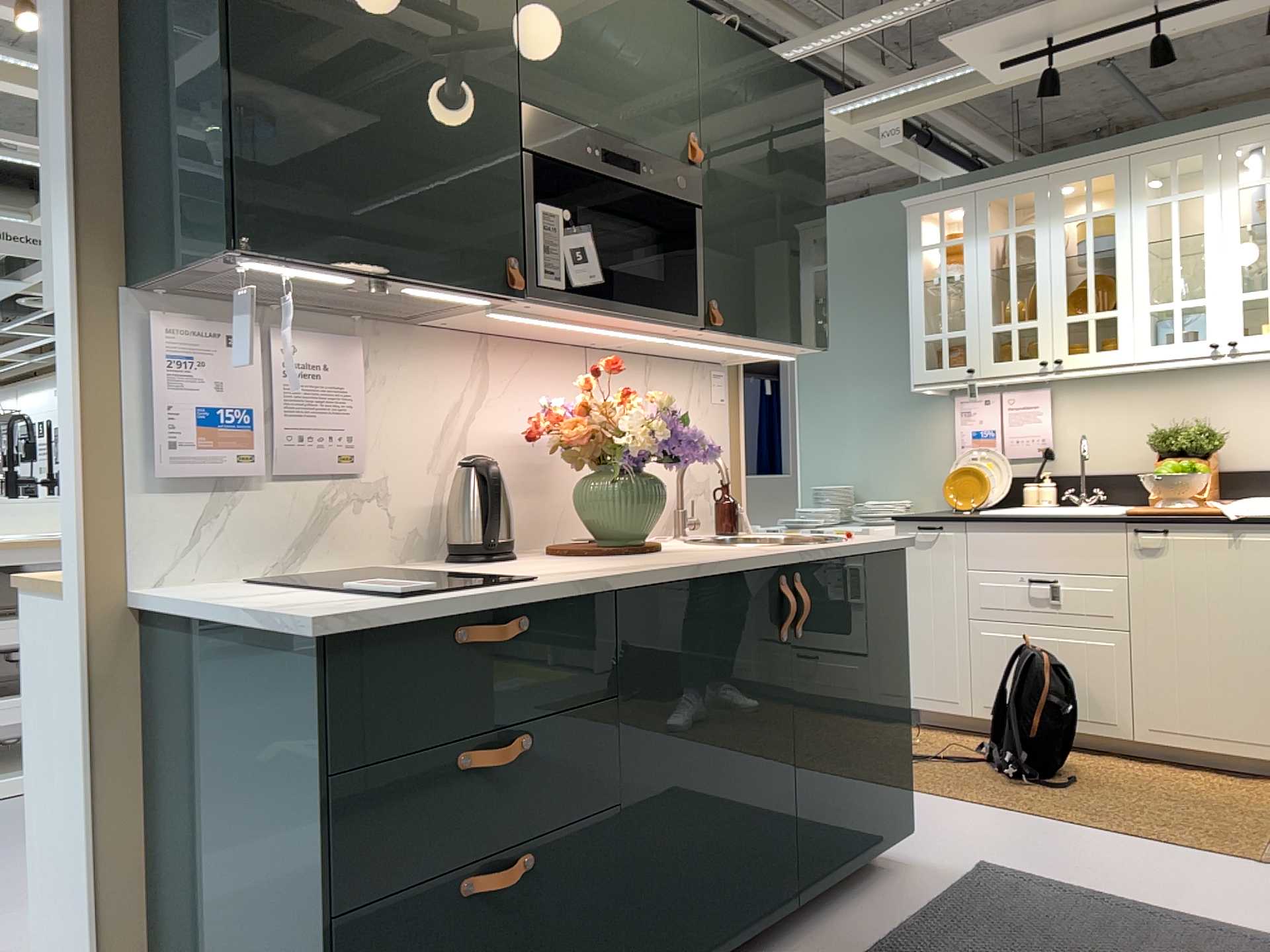} &
        \includegraphics[width=\sz\linewidth, height=\hw]{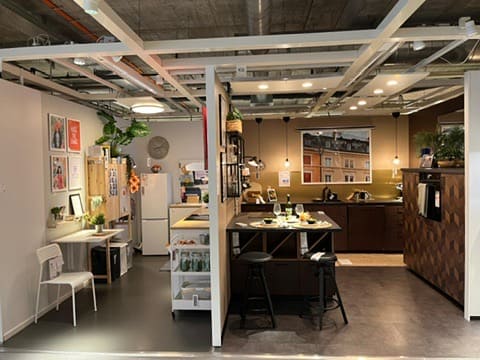} &
        \includegraphics[width=\sz\linewidth, height=\hw]{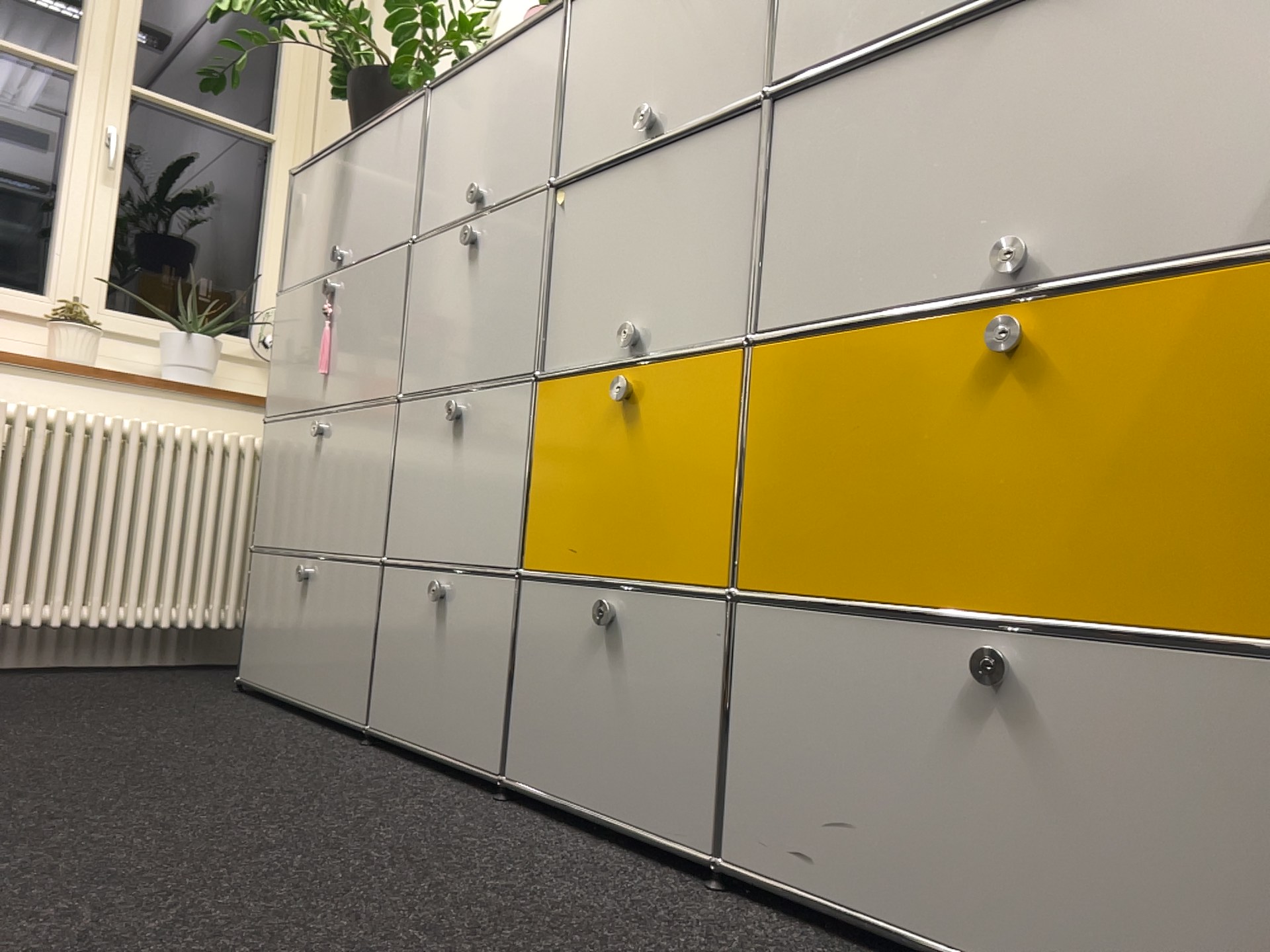} &
        \includegraphics[width=\sz\linewidth, height=\hw]{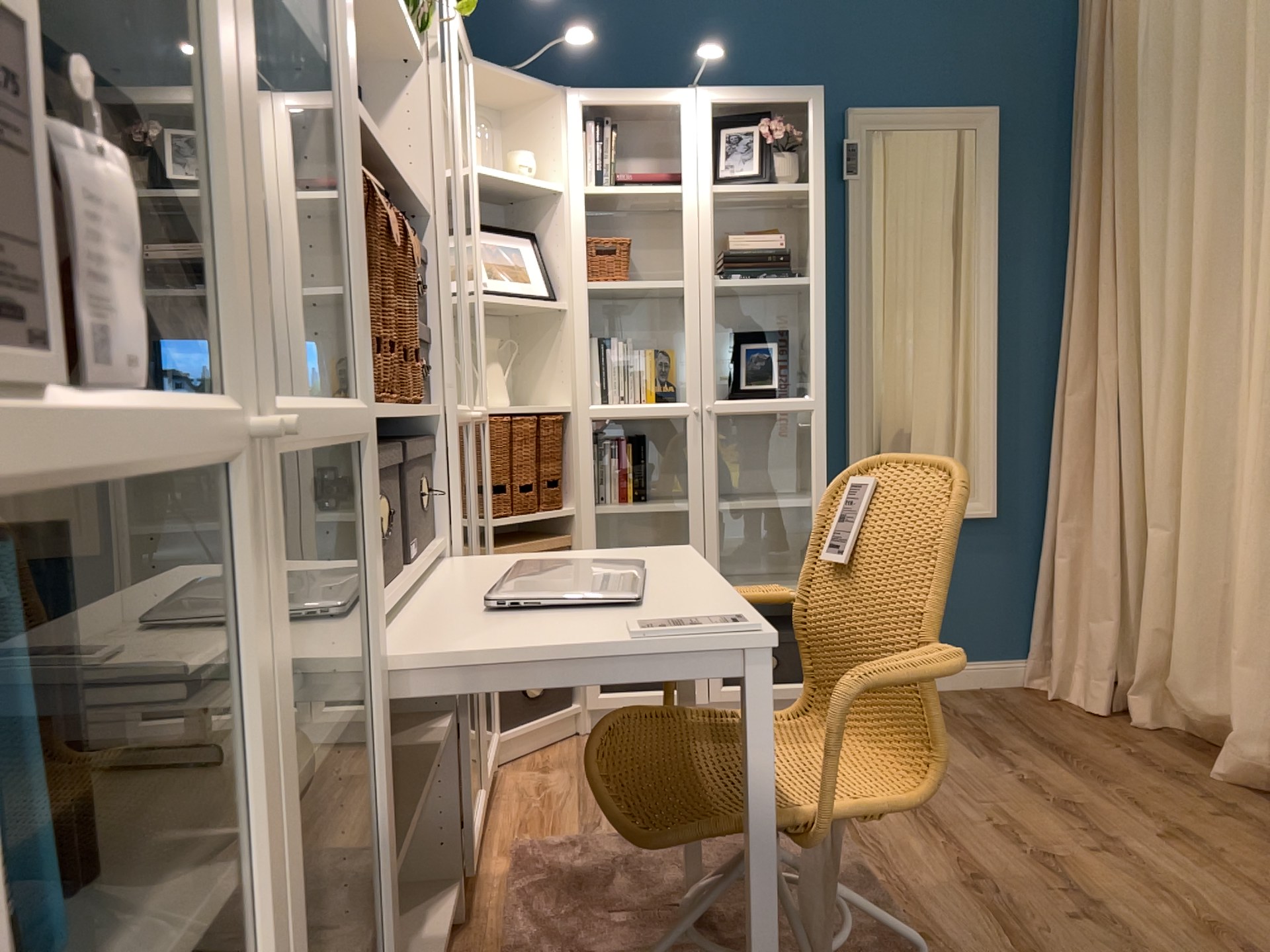} &
        \includegraphics[width=\sz\linewidth, height=\hw]{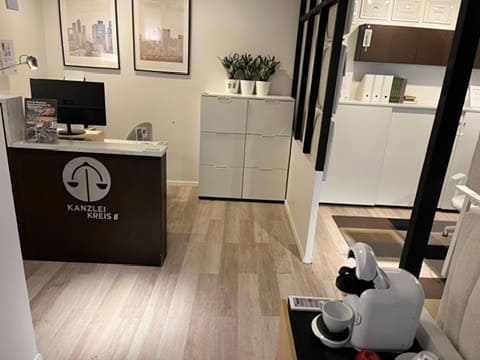} 
        \\
        \includegraphics[width=\sz\linewidth, height=\hw]{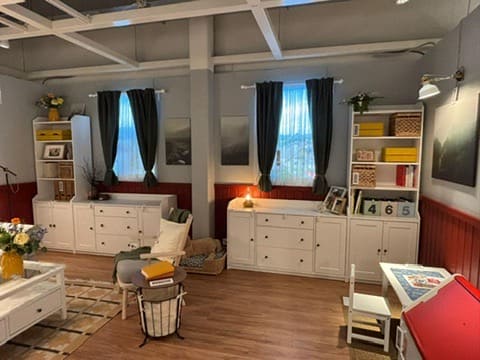} &
        \includegraphics[width=\sz\linewidth, height=\hw]{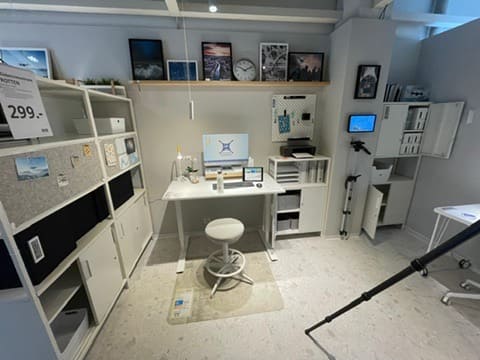} &
        \includegraphics[width=\sz\linewidth, height=\hw]{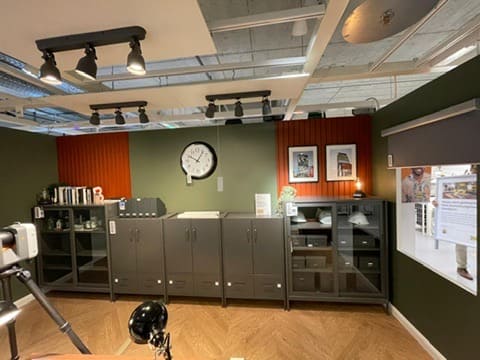} &
        \includegraphics[width=\sz\linewidth, height=\hw]{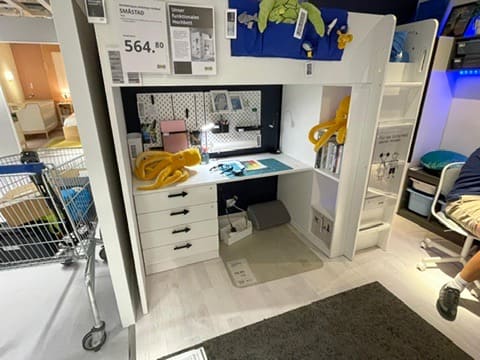} &
        \includegraphics[width=\sz\linewidth, height=\hw]{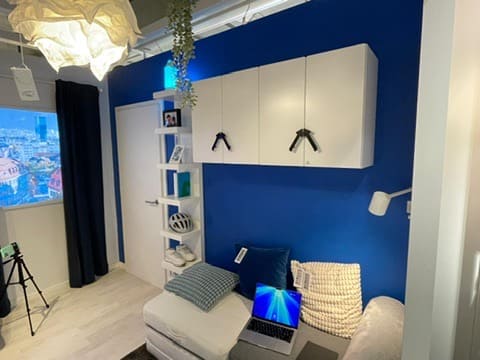} &
        \includegraphics[width=\sz\linewidth, height=\hw]{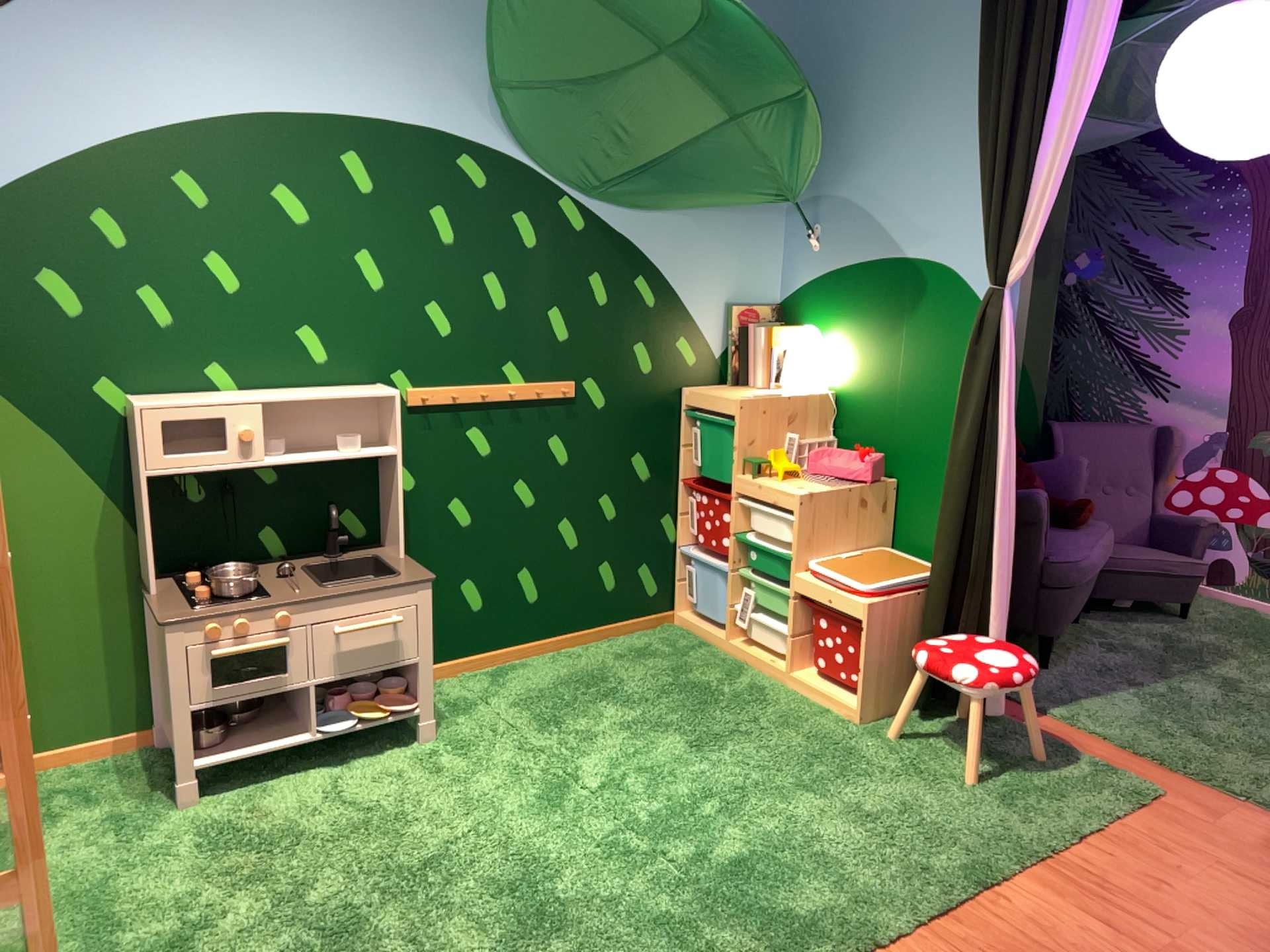} &
        \includegraphics[width=\sz\linewidth, height=\hw]{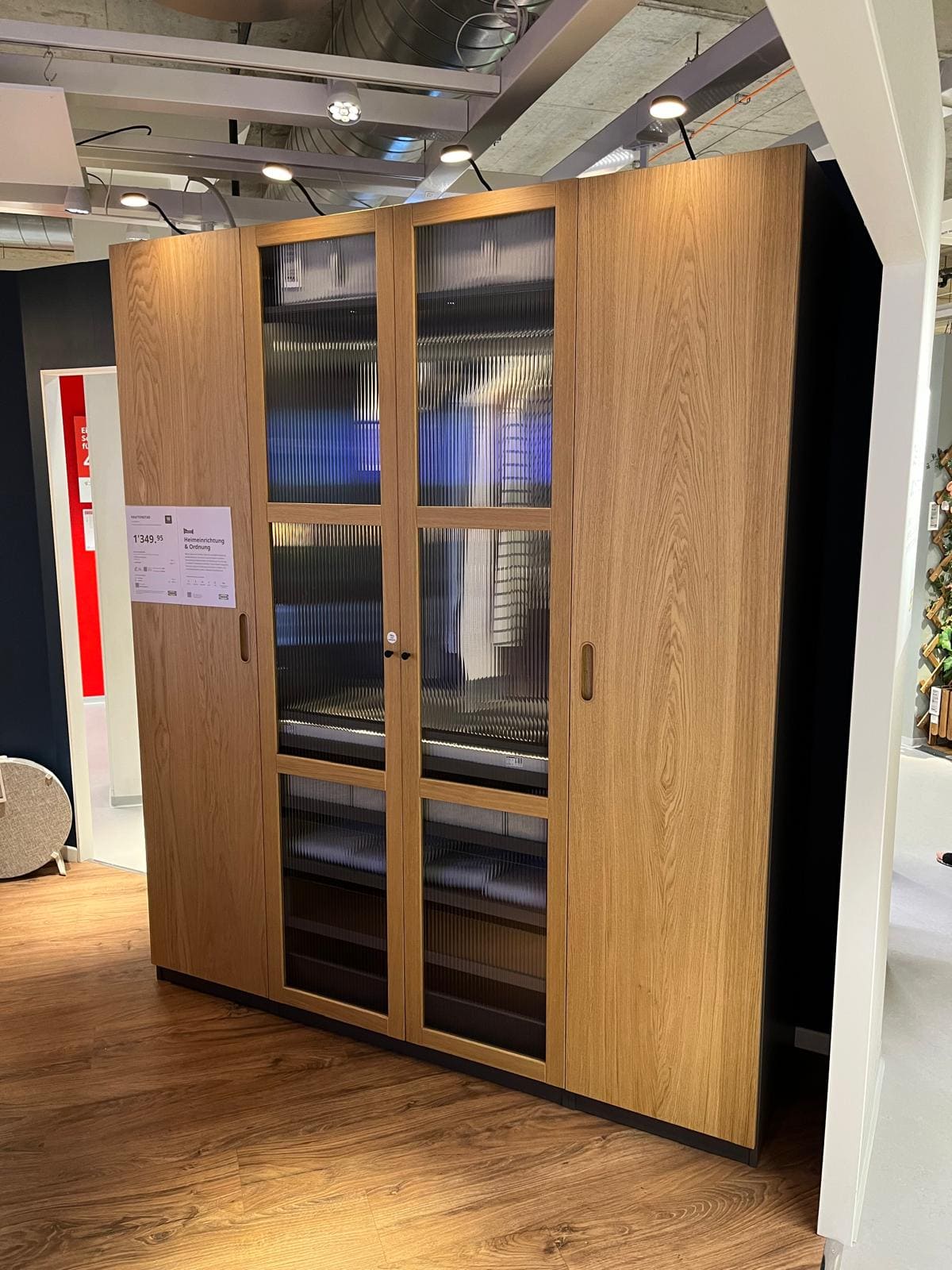} &
        \includegraphics[width=0.33\linewidth]{author-kit-CVPR2026-v1-latex-/figures/locations/Logo.png} 
        \\ \vspace{-3pt}
        
    \end{tabular}}
    \vspace{-10pt}
  \caption{\textbf{Locations of the Hoi! dataset.} A diverse collection of real-world indoor environments featuring kitchens, bathrooms, offices, and living spaces, where each has RGB-D sequences, GT, panoramic images, and various articulated objects that have interactions with multiple grippers and users.}
  \label{fig:locations}
  \vspace{-7pt}
\end{figure*}

\section{Related Work}


In the following, we analyze previous efforts in articulation and video understanding and delineate how our Hoi! dataset connects the fields of physical reasoning and perceptive interaction understanding.
\PAR{Articulation Understanding.} Early works modeled articulations as kinematic graphs and estimated joint parameters~\cite{sturm2013learning, urgenlearning}, but relied on known coordinate frames in controlled settings. Later efforts introduced large-scale simulated environments~\cite{xiang2020sapien, mo2019partnet} and real-world part databases~\cite{liu2022akb}.
However, these resources either remain fully simulated, limiting transfer to real manipulation, or provide object models without any interaction data.
Going beyond pure visual perception, datasets such as RBO~\cite{RBO_dataset} provide real RGB-D sequences of humans manipulating articulated objects and include limited force measurements, but remain small in scale and lack multi-view or multi-embodiment coverage. Learning-based approaches like FlowBot3D~\cite{eisner2022flowbot3d} estimate dense articulation flow fields for planning, yet rely largely on simulated training data. Complementary work on force prediction, such as object-centric models of everyday forces~\cite{Jain2013} and text-guided force reasoning for mobile manipulation~\cite{collins2023forcesighttextguidedmobilemanipulation}, demonstrates the value of modeling contact-force distributions for articulated tasks. Vision-only methods~\cite{song2024reactoreconstructingarticulatedobjects, heppert2023carto} and digital-twin approaches such as Ditto~\cite{jiang2022dittobuildingdigitaltwins} or ArtGS~\cite{liu2025artgsbuildinginteractablereplicas} focus on reconstructing articulated objects from video, but omit force or tactile interaction. In-the-wild articulation estimation has also emerged, with ArtiPoint~\cite{werby2025articulatedobjectestimationwild} inferring articulations from egocentric RGB-D. 
However, key gaps remain, as summarized in ~\cref{tab:overview_datasets}: (1) few public datasets combine articulated object manipulation with force/tactile sensing, (2) even fewer cover multiple viewpoints, e.g., ego, exo, wrist; and (3) almost none provide aligned recordings of both human and robotic embodiments performing the same articulated interactions across these modalities. As a consequence, it is difficult to systematically study how an action’s visual appearance relates to its physical patterns, or how human-demonstrated skills transfer to robot embodiments.
Our proposed dataset ``Hoi!" is explicitly designed to close this gap by jointly capturing vision, force, and tactile signals from humans and robots interacting with the same articulated objects, enabling research that links visual perception to haptic action and facilitates effective transfer of manipulation skills from human to robot.
\PAR{Video Understanding.}
Egocentric benchmarks such as \cite{EpicKitchen_Damen2022, grauman2022ego4d, grauman2024egoexo4dunderstandingskilledhuman, HDEpic_perrett2025hd} have enabled progress on action recognition, anticipation, and activity understanding in everyday environments. However, 
while they excel at depicting "what happened" in videos, they offer no information on the forces applied or contact feedback, making it hard to translate insights to the physical realm.
In robotics, recent efforts have leveraged large video collections to improve policy learning. \citet{nairr3m} learn universal visual representations from egocentric video, substantially accelerating downstream manipulation learning. This demonstrates that representations learned from human video can accelerate robot policy learning. Similarly, \cite{xiong2022robotube} and \cite{lepert2025phantomtrainingrobotsrobots} derive robot policies directly from human video demonstrations, eliminating the need for robot-collected data. \citet{kareer2025egomimic} accelerates imitation learning through easier-to-collect human demonstrations. These approaches illustrate the growing trend of leveraging human video data to teach robots, effectively narrowing the gap between computer vision and robotics. However, a crucial limitation is their limited generalization: Most of these works are restricted to tabletop settings and a low domain gap between demonstrations and deployment. Partially, this is because they strongly rely on visual inputs and omit the multimodal richness (like forces and tactile cues) that is central to physical manipulation. 
ForceMimic~\cite{liu2025forcemimicforcecentricimitationlearning} and RH20T~\cite{fang2023rh20tcomprehensiveroboticdataset} demonstrated that incorporating multimodal signals, particularly force measurements, can significantly improve robotic manipulation performance. 
As summarized in \cref{sec:dataset}, prior video datasets provide wide semantic coverage of human activity, but they do not provide the integrated multimodal sensing or cross-embodiment correspondence needed for grounded manipulation research.
In particular, without data that couples ``what is seen" with ``what is done" across both human and robotic embodiments, we cannot easily research an action's visual appearance correlated with the forces applied or how a skill demonstrated by a person might transfer to a robot's morphology. 
Motivated by this shortcoming, we jointly capture vision, force, and tactile signals across both human and robotic executions of identical tasks to bridge this gap. By providing aligned multimodal recordings 
, the dataset aims to enable research that tightly links visual perception to haptic action, ultimately facilitating effective learning and transfer of manipulation skills across embodiments.
\vspace{-5pt}

\section{Hoi! Dataset}

\label{sec:dataset}
\begin{table}[t]
\vspace{-10pt}
\centering
\setlength{\tabcolsep}{1pt}
\footnotesize
\resizebox{\linewidth}{!}{%
\begin{tabular}{llcccccccc}
\toprule
\multirow{2}{*}{\parbox{3cm}{\textbf{Manipulation}\\\textbf{Mode}}} &
\textbf{Gripper} &
\multicolumn{3}{c}{\textbf{Cameras}} &
\multicolumn{5}{c}{\textbf{Additional Modalities}} 
\\
\cmidrule(lr){3-5} \cmidrule(l){6-10}
& & $2\,\times$ Exo & Ego & Wrist &
\makecell{Wrist\\IMU} &
\makecell{Wrist\\Depth} &
\makecell{Force-\\Torque} &
\makecell{Finger\\Haptics} &
\makecell{Motor\\Torque}\\
\midrule
Hand only
& 5-finger
& \checkmark & \checkmark & \ding{55}
& \ding{55} & \ding{55} & \ding{55} & \ding{55} & \ding{55} \\
Hand + Wrist Cam
& 5-finger
& \checkmark & \checkmark & \checkmark
& \checkmark & \ding{55} & \ding{55} & \ding{55} & \ding{55} \\
UMI Gripper
& Antipodal
& \checkmark & \checkmark & \checkmark
& \checkmark & \ding{55} & \ding{55} & \ding{55} & \ding{55}  \\
Hoi! Gripper (ours)
& Antipodal
& \checkmark & \checkmark & \checkmark
& \checkmark & \checkmark & \checkmark & \checkmark & \checkmark\\
(Spot)
& Antipodal
& \checkmark & \ding{55} & \checkmark
& \ding{55} & \checkmark & \ding{55} & \ding{55} & \ding{55}\\
\bottomrule
\end{tabular}%
}
\vspace{-5pt}
\caption{\textbf{Recording setup.} Each manipulation condition comprises several recording modules producing multiple time-aligned data streams. \vspace{-15pt}}
\label{tab:recording-conditions}
\end{table}

\begin{figure}[t]
  \vspace*{-10pt}
  \centering
  \includegraphics[width=\columnwidth]{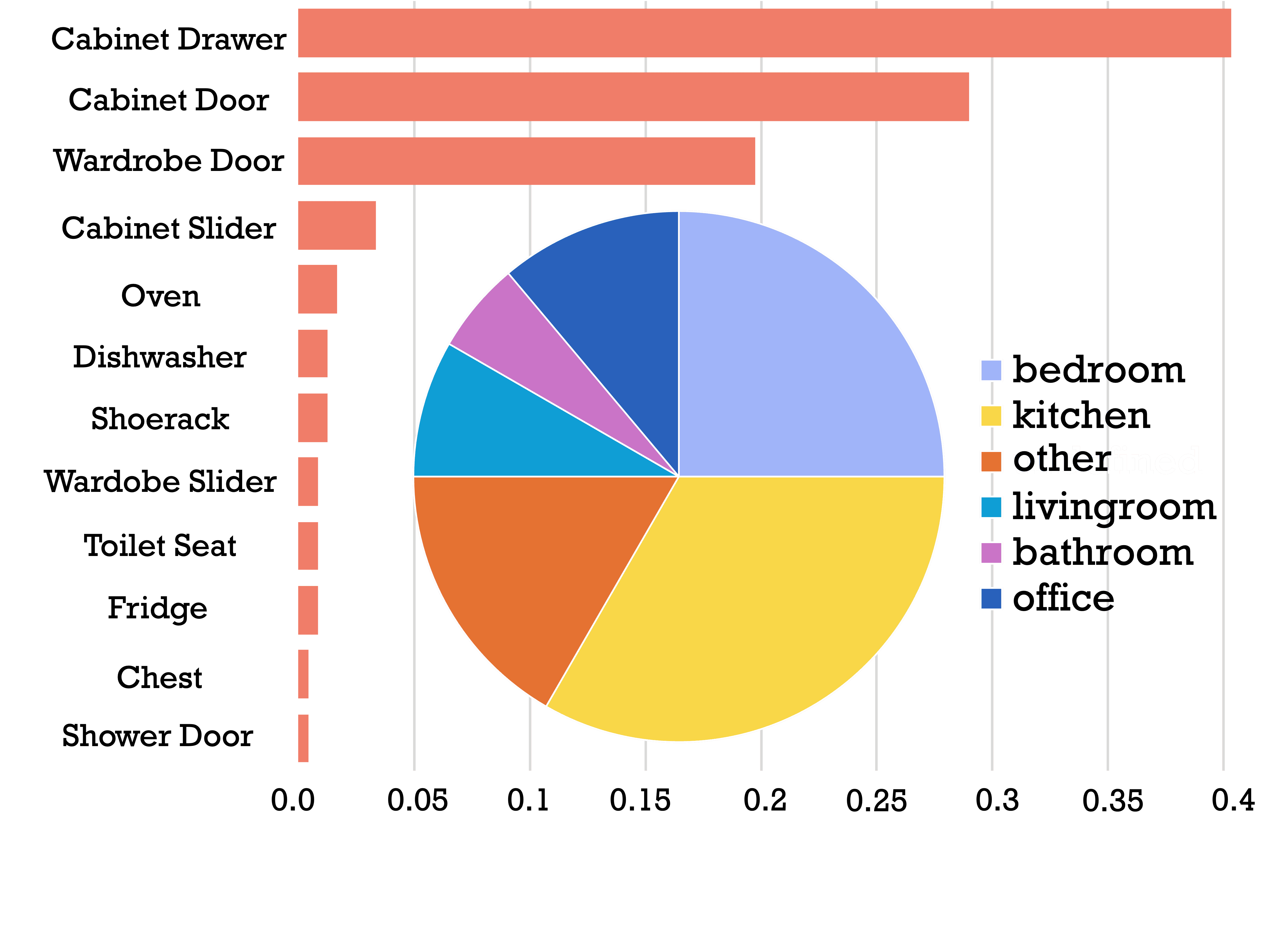}
  \vspace*{-40pt}
  \caption{\textbf{Distribution of environments and articulated interaction categories in the Hoi! dataset.}
The bar chart depicts the relative frequency of human interactions across articulated categories, while the inset pie chart summarizes the proportion of environments involved in the interactions.}
  \label{fig:dataset_stats}
    \smallskip
    \vspace*{-15pt}
\end{figure}

\begin{figure}[t]

  \centering

\vspace{-5pt}
\hspace*{-4pt}
\includegraphics[width=0.49\textwidth]{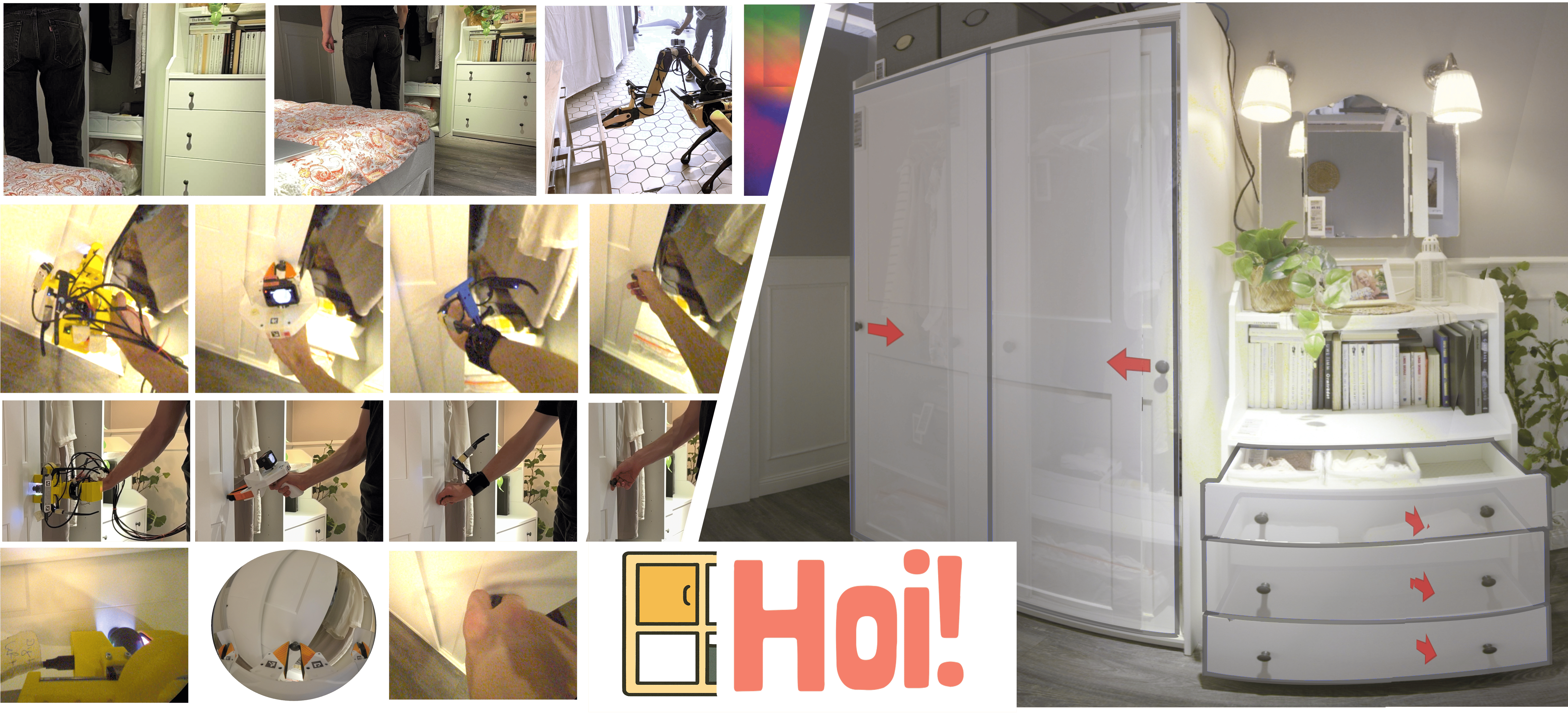}
  \caption{\textbf{Viewpoint recordings.} Recorded viewpoints for articulated part interactions. Each row corresponds to a different setup, showing synchronized exocentric (left), egocentric (center), and wrist-mounted (right) perspectives for both human and robot executions.}
  \vspace{-5pt}
  \label{fig:viewpoints}
\end{figure}

We introduce a dataset for force-grounded, cross-view, and cross-embodiment articulated manipulation. The dataset captures humans operating everyday articulated objects in realistic furnished rooms, with each interaction performed under four embodiments: (i) a human hand, (ii) a human hand with a wrist-mounted camera, (iii) a handheld UMI~\cite{chi2024universalmanipulationinterfaceinthewild} gripper, and (iv) a custom Hoi! gripper equipped with force–torque and tactile sensing. We record a small subset of the interactions using a teleoperated Spot robot, equipped with body cameras and a wrist-mounted Aria. An overview of the articulated parts is shown in~\cref{fig:dataset_stats}. Across all embodiments, interactions are recorded from multiple viewpoints: an egocentric camera (Project Aria~\cite{engel2023projectarianewtool}) providing RGB, SLAM, eye gaze, and hand pose, and two static third-person views captured with iPhone 13 Pro devices (RGB + LiDAR depth). \cref{tab:recording-conditions} summarizes the data streams under each condition, and~\cref{fig:viewpoints} shows all viewpoints.
All modalities are temporally and spatially aligned, enabling direct comparison of how the same articulated object is operated across different embodiments. To anchor these interactions in accurate geometry, we additionally capture before/after high-resolution 3D scans of each environment using a Leica RTC360 laser scanner. This fully aligned and multimodal setup creates a unique foundation for studying how visual observations relate to physical interaction forces across human and robotic embodiments.

\begin{figure}[t]
\vspace{-18pt}
    \centering
\includegraphics[width=0.8\linewidth]{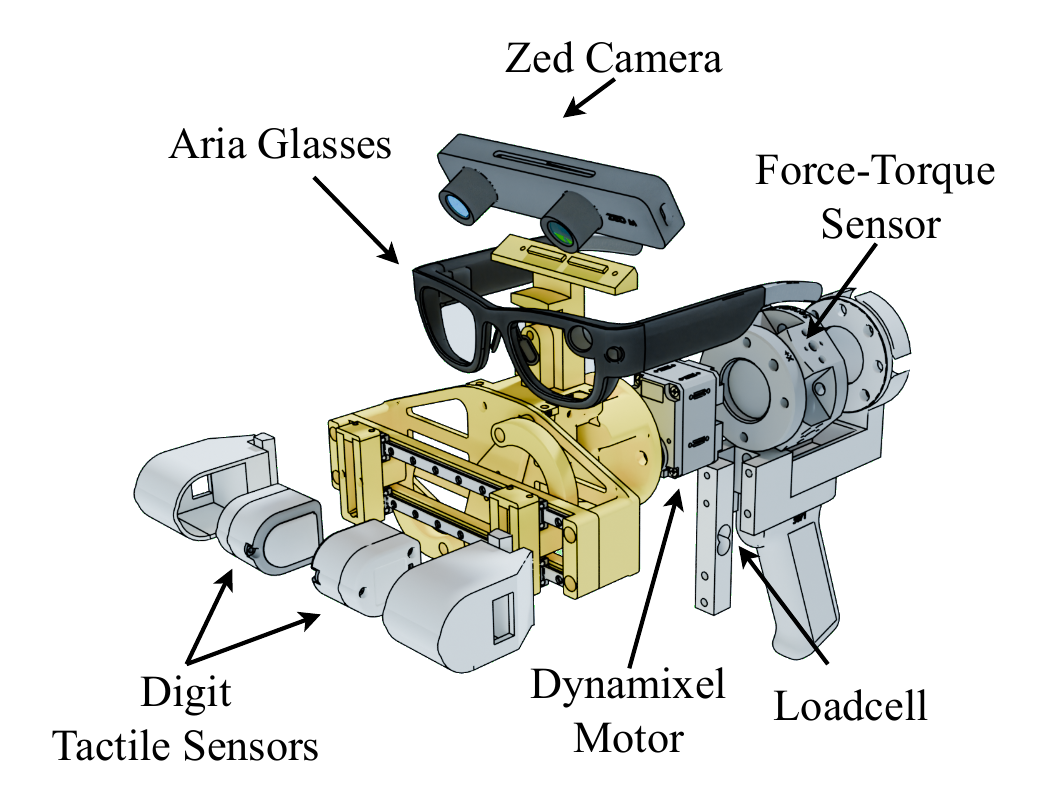}\vspace{-8pt}
    \caption{\textbf{Hoi! Gripper.} The 2-finger parallel gripper is operated through the load cell, where the measured load is translated into gripping force. Interaction force and tactile contact pressure are measured through the Digit and Force-Torque sensors respectively. Aria Glasses and a stereo camera provide pose estimation and wrist-view observations. We will release the design as open source.}
    \label{fig:gripper}
    \smallskip
    \vspace*{-17pt}
\end{figure}

\begin{figure}[t]
\vspace*{-5pt}
\hspace*{-20pt}
\includegraphics[width=1.1\linewidth]{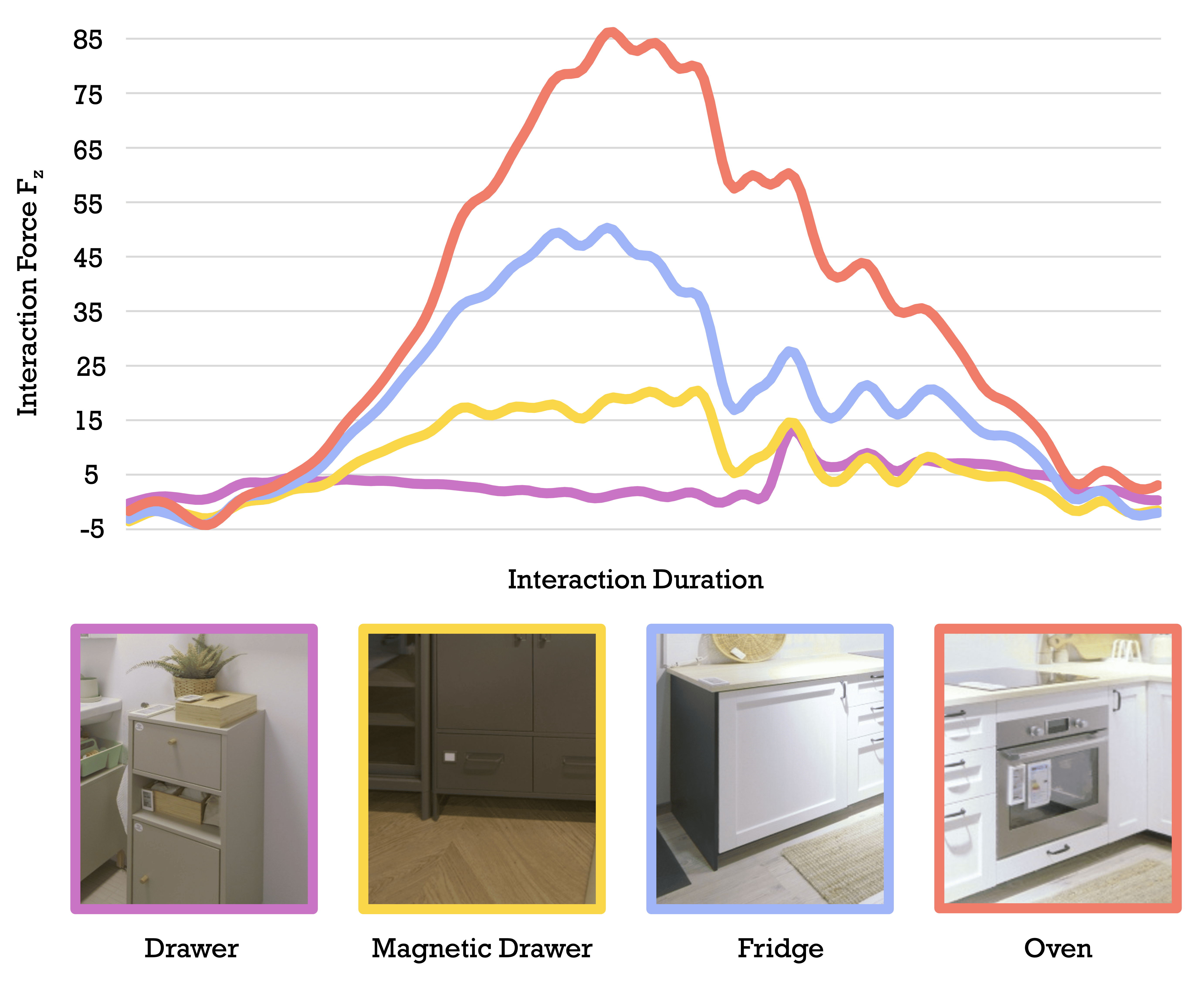}
\caption{\textbf{Example of the measured interaction forces for several articulated elements}. Each curve corresponds to a different component (highlighted in matching colors below), illustrating how force magnitudes vary across types of articulated parts.}
    \smallskip
\label{fig:interaction_forces}
\end{figure}

\PAR{Hoi! Gripper.} We design and open-source a custom gripper (\cref{fig:gripper}) to capture high-quality tactile and force data during interaction. The gripper features two opposing GelSight Digit sensors for high-resolution tactile imaging and an antipodal mechanism inspired by the ALOHA design~\cite{zhao2023learningfinegrainedbimanualmanipulation}. The gripper is powered by a Dynamixel XM430-W350-T motor and features a Bota SensONE 6-DoF force–torque sensor to measure interaction forces at the wrist.
To support human-operated demonstrations, the assembly is mounted on a handheld “gripper-on-a-stick,” with gripping force modulated via a calibrated load cell in the handle. A wrist-mounted ZED Mini stereo camera and Project Aria device provide pose tracking and RGB-D wrist observations.
This setup allows us to couple visual observations with precise force–torque and tactile signals, providing a rich basis for investigating real articulated interactions. As illustrated in \cref{fig:interaction_forces}, force profiles vary substantially across different articulated objects, highlighting how force signals offer complementary information beyond visual cues.
The full system runs on a battery-powered NVIDIA Jetson Orin Nano carried in a backpack, enabling fully mobile data collection. The sensor suite is fully calibrated and the gripper is gravity-compensated. Detailed specifications are provided in the supplementary.



\begin{figure*}[t]
    \vspace*{-21pt}
  \centering
  \includegraphics[width=\textwidth]{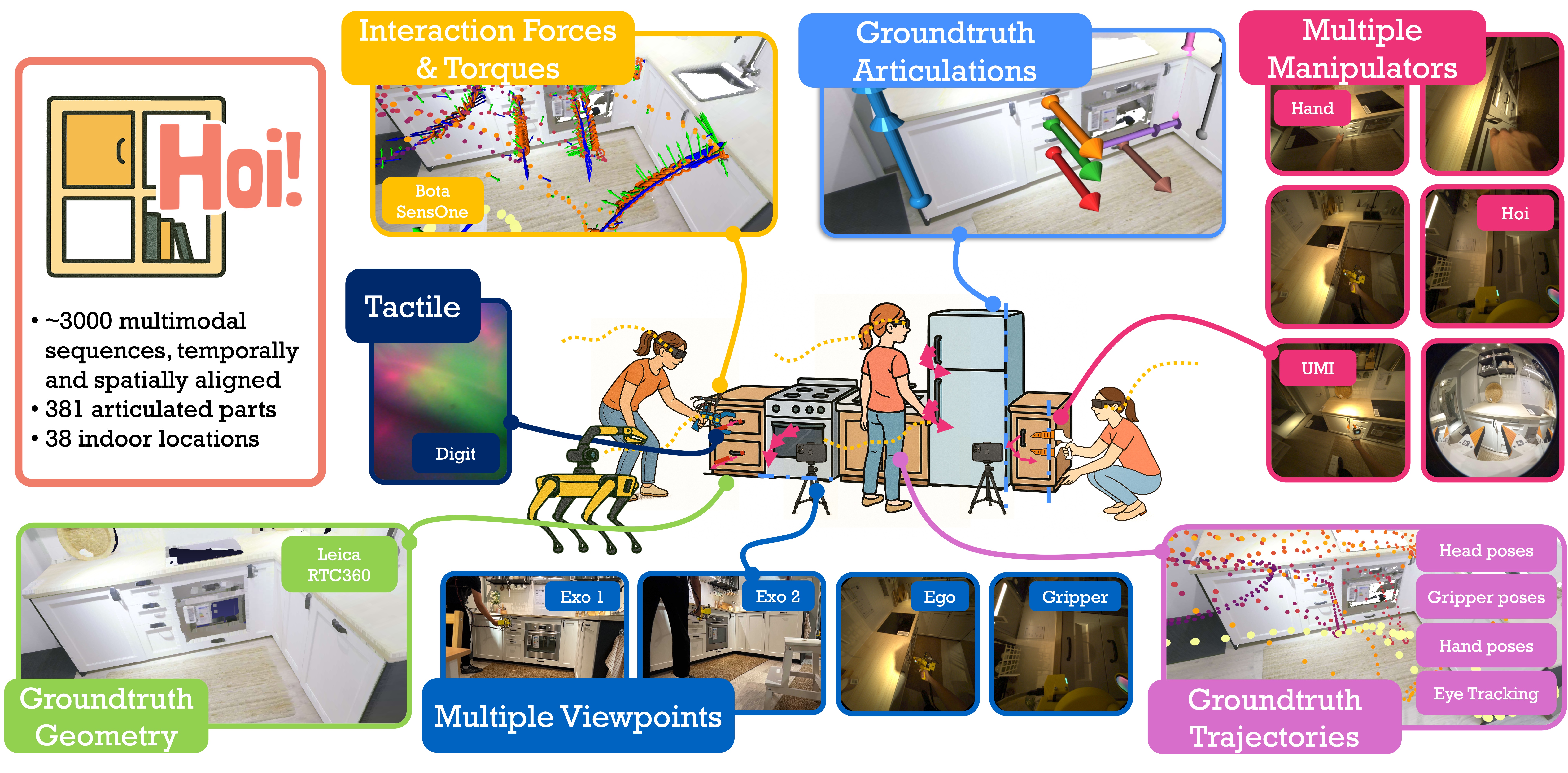}
\vspace*{-23pt}
  \caption{\textbf{Overview of the dataset collection setup.} The dataset consists of \textbf{3048} multimodal sequences capturing human interactions with \textbf{381} articulated objects across \textbf{38} locations using multiple viewpoints (egocentric and third-person cameras) and manipulation conditions (hand, gripper-based). Ground truth data includes trajectories, contacts, haptic feedback, force measurements, and high-resolution 3D point clouds of each environment.}
  \label{fig:dataset_overview}
\vspace*{-14pt}
\end{figure*}

\PAR{Data Collection.} We record the dataset primarily in the exhibition area of a furniture store, with additional sequences captured in a university lab and private apartments. 
A team of seven human demonstrators performs all interactions, each under the four manipulation conditions described above and in ~\cref{tab:recording-conditions}.
To ensure consistent capture and simplify post-processing, we follow a structured collection protocol. Each recording session covers between 3 and 11 articulated parts using a single manipulation condition (e.g., hand only). Once all devices start recording, we present a dynamic QR code encoding the current timestamp to every video stream. We then mark the start of each individual interaction by briefly showing a static QR code to one exocentric camera. The operator proceeds through all articulated parts in alternating open/close order. 
For each environment, we also capture high-resolution 3D point clouds with a Leica RTC360 laser scanner. We first scan the unarticulated scene, followed by scans in which as many articulated parts as possible are opened without occluding each other, resulting in 2-5 point clouds per location.
These scans provide ground-truth geometry for all environments and articulations and serve as a shared reference frame for spatial calibration across devices.

\PAR{Time Alignment.} Since the different recording modules run on independent internal clocks, we align all streams in post-processing. During recording, we display a QR code encoding the current Unix timestamp at $25\,\text{Hz}$ into each camera stream $i$. Detecting and decoding this QR code yields a time offset for each video stream with respect to a common reference time, allowing us to express all modalities on a shared timeline. For typical frame rates of $30$–$60\,\text{Hz}$, this procedure results in a temporal alignment accuracy of ca. $10$–$25\,\text{ms}$. 

\PAR{Pose Data.} We use Aria Machine Perception Services (MPS) to generate 6-DoF device poses as well as hand poses and eye gaze data from the egocentric Aria video stream. 
For the wrist camera and the Hoi! Gripper we also use the attached Aria devices and extrinsic calibrations to pose all sensors.
For the UMI gripper, we follow their setup and use ORB-SLAM3~\cite{campose2021orb} to track the gripper-mounted GoPro camera and globally refine the trajectory using GTSAM~\cite{gtsam}.
Because the third-person iPhone cameras remain static, we can directly measure their poses in a common reference frame during spatial calibration, as described next.

\PAR{Spatial Alignment.}
We spatially align all recording devices into a common reference frame using visual localization against the high-resolution 3D scans. We first construct a reference set of dense 2D-3D correspondences from the point cloud and corresponding panoramic images captured with the scanner. 
We assume zero drift and global consistency of the per-device SLAM, which reduces the global registration problem to a single rigid 3D-3D alignment. 
We automatically select a set of high-quality keyframes for robust registration. 
We estimate the 6-DoF pose of each keyframe
through hloc~\cite{sarlin2019coarse} with the 2D-3D database constructed from the Leica scan and robustly estimate a single rigid transformation $\mathbf{T}_{\text{world}}^{\text{query}}$ between each sensor trajectory and the shared world frame.
\PAR{Alignment Accuracy.}
We evaluate trajectory accuracy using a Qualisys motion capture (mocap) system. Mocap markers attached to the Aria frame provide the ground-truth trajectory, while the Aria MPS gives the corresponding device trajectory. The rigid transform between the mocap body frame and the Aria device frame is obtained via hand–eye calibration~\cite{Furrer2017FSR}. Using this calibration, we express both trajectories in a common frame, align them and directly compare them for evaluation. Trajectory errors are listed in ~\cref{tab:aria_traj_error}.
\begin{table}[t]
  \centering
  \caption{\textbf{Trajectory Evaluation.} Trajectory error of Aria-derived quantities, evaluated against Qualisys motion-capture trajectories.}
  \label{tab:aria_traj_error}
  \vspace{-5pt}
  \setlength{\tabcolsep}{7pt}
  \footnotesize
  \begin{tabular}{lccc}
    \toprule
    \textbf{Metric} & \textbf{Head} & \textbf{Wrist} & \textbf{Gripper} \\
    \midrule
    RMSE position [m] $\downarrow$     & 0.005 & 0.005 & 0.006 \\
    RMSE rotation [rad] $\downarrow$   & 0.016 & 0.012 & 0.012 \\
    \midrule
    Acc@1cm, 1$^\circ$ $\uparrow$      & 0.810 & 0.886 & 0.810 \\
    Acc@5cm, 5$^\circ$ $\uparrow$      & 0.998 & 1.000 & 1.000 \\
    Acc@10cm, 10$^\circ$ $\uparrow$    & 1.000 & 1.000 & 1.000 \\
    \bottomrule
  \end{tabular}
  \vspace{-17pt}
\end{table}

\PAR{Annotations.} For each articulated object, we gather all individual interaction recordings based on automatic video cutting with the QR code as described above. We then manually verify using a light-weight annotation tool. In addition, we use the tool of~\cite{werby2025articulatedobjectestimationwild} to annotate the articulation type (\texttt{prismatic} or \texttt{revolute}) and axis. We further extend the annotation tool to add a 3D mask of the object part and a language description of the part. To generate the mask, we prompt SAMv2~\cite{ravi2024sam2segmentimages} on the panoramic images and lift the predicted mask to 3D using the point cloud.
\section{Evaluations}
\label{sec:evaluations}

Transferring manipulation skills across different embodiments, for instance from a human hand demonstration to a robot gripper, requires a fundamental understanding of what interactions each object affords and the object's physical properties. 
An agent must deduce how an object can be manipulated (\eg does a handle pull outwards or twist, does a door swing left or right) and recognize constraints that dictate the required forces or motions, \eg, a drawer might be heavier or latched, requiring more force to open). 
Our evaluation examines embodied object understanding via three complementary tasks with corresponding benchmarks: articulated object estimation (perceiving how parts move and estimating their kinematics) in Sec.~\ref{sec:articulated-pose-estimation}, tactile force estimation (inferring contact forces from touch) in Sec.~\ref{sec:force-estimation}, and visual force estimation (predicting the force needed to achieve a desired interaction from visual input) in Sec.~\ref{sec:visual-force-estimation}. 


\begin{table*}[t]
\vspace{-15pt}
    \centering
    \footnotesize
    \resizebox{0.7\linewidth}{!}{%
    \begin{tabular}{llccccccc}
    \toprule
        Dataset & Method & \multicolumn{2}{c}{Articulation Type} & \multicolumn{5}{c}{Motion Parameters}\\
        \cmidrule(lr){3-4} \cmidrule(lr){5-9} 
        &  & $R_\textrm{prismatic}$ [\%] & $R_\textrm{revolute}$ [\%] & $\theta_{err}^{pris}$[deg] & $\theta_{err}^{rev}$[deg] & $d_{L2}$[m] \\
        \midrule
               & GPT-5 (egocentric) & 72.4 & 78.8 & - & - & - \\
        Arti4D~\cite{ARTI4D_werby2025} & ArtGS & 100 & 0 & 52.59 & 56.82 & 0.25\\
               & ArtiPoint~\cite{ARTI4D_werby2025} & 68 & 98 & 14.54 & 17.14 & 0.07\\

        \midrule
        Hoi! (ours) & GPT-5 (egocentric) & 71.9 & 89.7 & - & - & - \\
       &  GPT-5 (exocentric) & 65.6 & 89.7 & - & - & - \\
        & ArtGS~\cite{liu2025artgsbuildinginteractablereplicas} & 100.0 & 0.00 & 58.39 & 49.11 & 0.321 \\
        & ArtiPoint~\cite{ARTI4D_werby2025} & 26.90 & 57.10 & 47.06 & 63.76 & 0.540 \\
        \bottomrule
    \end{tabular}}
    \vspace{-7pt}
    \caption{\textbf{Articulation Estimation.} Given a single image before interaction (for GPT) or the egocentric video (for ArtGS and ArtiPoint), methods estimate the type of articulation as well as the exact articulation axis in 3D. We report type recall $R$, axis angle error $\theta_{err}$, and distance error for revolutes $d_{L2}$}
    \label{tab:articulation-prediction}
    \vspace{-11pt}
\end{table*}

\subsection{In-the-Wild Articulated Object Estimation}
\label{sec:articulated-pose-estimation}

We consider articulation understanding as a key prerequisite for manipulation: before an agent can reason about the forces required to act on an object, it must first infer how its motion is governed. Our dataset enables this by providing posed RGB observations (egocentric and exocentric), 3D scene reconstructions, and manually annotated articulation parameters, allowing us to evaluate articulation estimation methods under realistic in-the-wild conditions.
We evaluate three representative approaches: ArtiPoint~\cite{werby2025articulatedobjectestimationwild} and the Gaussian-Splatting-based ArtGS~\cite{liu2025artgsbuildinginteractablereplicas} predict both articulation type and motion parameters. In addition, we evaluate GPT-5 as a VLM to infer articulation types directly from single RGB views - both egocentric and third-person.
Since Aria glasses do not provide dense depth, we generate depth inputs using MapAnything~\cite{keetha2025mapanythinguniversalfeedforwardmetric}. Predicted depth maps are aligned to metric scale by rendering depth from our 3D scene meshes at the corresponding camera pose and estimating a global scale factor via the mode of the per-pixel depth-ratio histogram. This compensates for regions where predicted depth and rendered geometry differ (\eg, hands or moving articulated parts). More details are included in the supplementary material.
We observe that methods such as ArtiPoint and ArtGS exhibit significantly lower performance on our dataset compared to Arti4D~\cite{ARTI4D_werby2025} as indicated in ~\cref{tab:articulation-prediction}. ArtGS exhibits lower performance in in-the-wild settings due to clutter and non-robust object segmentation on both Arti4D and Hoi!. However, ArtiPoint performs worse on Hoi! when faced with scaled monocular depth, which yields stochastic non-steady noise throughout interactions, severely limiting its 3D lifting and trajectory filtering. In addition, we find that mere articulation type prediction is surprisingly robust when relying on GPT-5 on both Hoi! and Arti4D. We conclude that current articulation estimation methods are either too dependent on accurate depth or fail in the presence of clutter and hands.

\subsection{Tactile Force Estimation}
\label{sec:force-estimation}
We investigate force prediction from gel-based tactile images using data collected with our dataset. The goal is to estimate the normal and tangential forces acting on the gripper solely from the tactile images produced by the Hoi! gripper’s GelSight Digit sensors during interaction. This task directly reflects the multimodality enabled by our dataset: tactile images from the Digits are temporally aligned with the corresponding end-effector forces from the force–torque (FT) sensor and the gripper's motor-induced gripping forces, allowing us to construct reliable ground-truth labels.
We evaluate two versions of the Sparsh model~\cite{higuera2024sparsh}, the state-of-the-art method for self-supervised tactile representation learning, using both the DINO~\cite{caron2021dino} and DINOv2~\cite{oquab2024dinov2} decoders with the force-estimation head (“Task~1”) from the original work.
To obtain ground-truth force components, we decompose interaction forces into normal and tangential directions and express all measurements in a common interaction frame aligned with the Digit sensors. External forces measured by the FT sensor are rotated into this frame, while the gripper’s internal gripping force is estimated from the torque–current relationship, its Jacobian, and a load-dependent calibration factor. We aggregate left/right sensor contributions to obtain combined normal, tangential, and total force magnitudes.
Since Sparsh is trained on forces within a known range, we clip the combined force magnitudes to remain consistent with the model’s expected distribution, avoiding evaluation-time extrapolation. Full details are provided in the supplementary material.
We report the RMSE of the estimated forces in~\cref{tab:force-rmse-ci-compact} and observe errors on the order of several Newtons, whereas Sparsh achieves millinewton-level accuracy on its original benchmark. Although a direct comparison is not fully equivalent - our setup evaluates forces aggregated across two opposing Digit sensors - this large increase in error is noteworthy. We attribute this degradation primarily to out-of-distribution contact geometries (the model was trained on a small set of simple indenters, whereas real handles, edges, and furniture parts present far more complex contact shapes) and to out-of-distribution load regimes that naturally arise during in-the-wild human operation. This highlights that tactile models which excel in controlled lab settings struggle to generalize to unconstrained real-world interactions, highlighting the value our data have for future research.

\begin{table}[t]
  \centering
  \footnotesize
  \caption{\textbf{Interaction Force Prediction.} Based on measurements from the DIGIT tactile pressure sensor. 
  RMSE (95\% CI) in Newtons, averaged over all validation environments.}
  \label{tab:force-rmse-ci-compact}
\vspace{-4pt}
  \setlength{\tabcolsep}{5pt}
    \resizebox{1.0\linewidth}{!}{%
\begin{tabular}{lccc}
\toprule
 Method & Tangential & Normal & Combined\\
\midrule
Sparsh\cite{higuera2024sparsh} w/ DINO & 3.07\tiny{[2.87, 3.26]} & 3.45\tiny{[3.24, 3.66]} & 3.86\tiny{[3.62, 4.11]}\\
 Sparsh\cite{higuera2024sparsh} w/ DINOv2 & 3.18\tiny{[2.99, 3.38]} & 3.79\tiny{[3.61, 3.96]} & 4.11\tiny{[3.90, 4.33]}\\
\bottomrule
\end{tabular}
}
\vspace{-10pt}
\end{table}

\subsection{Visual Force Estimation}
\label{sec:visual-force-estimation}
We evaluate the utility of our dataset for visual force estimation, where a model predicts a 3D interaction force (and affordance) from an RGB-D observation given a manipulation goal. For example, an RGB-D image of a drawer together with “open the drawer” should yield where to interact and what force to apply. We benchmark the ForceSight model~\cite{collins2023forcesighttextguidedmobilemanipulation}, which predicts force goals for text-guided manipulation and has shown that such goals can improve robotic performance.
Our dataset provides the necessary multimodality for this task: the Hoi! gripper supplies per-image force–torque measurements, each sequence includes a language goal derived from our annotations, and 3D ground-truth trajectories allow accurate alignment across frames. 
Because raw force–torque readings may include operator-induced forces unrelated to the actual articulation, we report results on both the raw signals and a motion-aligned version. We project measured forces and torques onto the gripper’s linear and rotational velocity directions, removing components that do not align with the intended motion.
Following the evaluation protocol of the original ForceSight paper, we evaluate the method in a zero-shot setting on our dataset. 
While the model achieves an RMSE of $0.404\,\mathrm{N}$ on the original dataset, its performance degrades noticeably when applied to our data. As shown in~\cref{tab:forcesight_rmse}, it particularly struggles in locations containing multiple articulated objects that require higher operating forces (\eg \textit{kitchen\_7} with an RMSE of $3.531\,\mathrm{N}$, which includes a fridge and an oven, or \textit{office\_1} with an RMSE of $2.325\,\mathrm{N}$, which features magnetic drawers; see~\cref{fig:interaction_forces}). This suggests limited prior exposure to stiff or force-demanding articulated mechanisms, an underrepresented class in existing datasets and prior work. We also observe a noticeable improvement when using the projected forces and torques, indicating that the method is less robust to real-world operation disturbances present in the raw measurements. Again, we highlight the value of our dataset to this underexplored line of research.

\begin{table}[h!]
\centering
\caption{\textbf{Visual Force Estimation.} Given an RGB-D observation and a manipulation goal (e.g., “open the drawer”), the model predicts the 3D interaction force required to perform the action. We report the force RMSE (in N, lower is better) of ForceSight \cite{collins2023forcesighttextguidedmobilemanipulation} across different locations in our dataset. \emph{Projected} denotes evaluation on force components aligned with the gripper’s motion direction.}

\vspace{-4pt}
\resizebox{\linewidth}{!}{%
\begin{tabular}{l c c}
\toprule
\textbf{Location} & \textbf{RMSE Projected [N]} & \textbf{RMSE Raw [N]} \\
\midrule
bathroom\_2   & 1.21 & 2.09 \\
bedroom\_4    & 1.33 & 1.85 \\
bedroom\_6    & 2.10 & 2.43 \\
kitchen\_7    & 3.53 & 3.64 \\
office\_1     & 2.33 & 3.69 \\
livingroom\_1 & 1.09 & 1.74 \\
\cmidrule(lr){2-3} 
Hoi! (Ours) & 2.23 & 2.57 \\  \midrule
ForceSight Dataset & -- & 0.40 \\
\bottomrule
\end{tabular}}
\vspace{-15pt}
\label{tab:forcesight_rmse}
\end{table}

\section{Limitations \& Future Work}
\label{sec:limitations}
Hoi! represents a first step toward bridging human and robot embodiments through force-grounded, cross-view interaction data, but several limitations remain. 
First, although human demonstrations with the Hoi! gripper mimic robotic end-effector interactions, they are still generated by a human operator and therefore do not fully capture the kinematic and dynamic constraints of real manipulators. This hybrid embodiment simplifies skill transfer in some way but complicates it in others since true generalization across full-body morphology is still an open challenge. Second, while our dataset spans a wide variety of articulated household objects, it does not yet cover the full range of mechanical complexities or rare edge-case mechanisms found in real-world environments. Finally, our current benchmarks are object-centric and focus on perception tasks such as force estimation and articulation reasoning. These serve as foundational capabilities, but extending them toward end-to-end policy learning, where perception and action are trained jointly, remains an important direction for future work.

\section{Conclusions}
We present Hoi!, a multimodal dataset designed to bridge the longstanding gap between human-centric and robot-centric interaction data. Central to our approach is the Hoi! gripper, which allows human operators to produce natural demonstrations with robot-grade physical sensing, capturing force, torque, tactile, and visual signals in real-world manipulation tasks. This setup enables a unified representation of interaction that is transferable across embodiments. Our evaluation shifts focus towards an object-centric view of manipulation: we assess how well agents can infer what an object affords, how it moves, and what physical effort is required to interact with it. Across three tasks: tactile force estimation, articulated motion prediction, and visual force estimation, we demonstrate how understanding object constraints is essential for generalizable manipulation and how current methods show significant room for improvement. By contributing a richly annotated, cross-embodiment and cross-view dataset, we aim to support the development of agents that not only observe but truly understand and act upon the physical world.

\newpage
\section{Funding \& Acknowledgments}
This research was partially funded by the ETH AI Center, ETH Foundation Project 2025-FS-352, the SNSF Advanced Grant 216260, and the Lamarr Institute for Machine Learning and Artificial Intelligence. It was further supported by the Robotics Institute Germany, Google, and Meta. We want to specifically thank the Meta ARIA team around James Ford, Jakob Engel, Mingfei Yan and Richard Newcombe for their support and valuable inputs.
We sincerely thank Tifanny Portela for her valuable support during data acquisition.

{
    \small
    \bibliographystyle{ieeenat_fullname}
    \bibliography{main}
}

\clearpage
\setcounter{page}{1}
\maketitlesupplementary
\appendix
\renewcommand{\figurename}{Supplementary Figure}
\renewcommand{\tablename}{Supplementary Table}
\begin{center}
\begin{minipage}{0.48\textwidth}  
\setlength{\parskip}{0pt}         
\makeatletter
\makeatother
{\small                
\tableofcontents
}
\end{minipage}
\end{center}
\vspace{5pt}

\section{Hoi! Dataset Details \& Gripper Calibration Details}
In the following we give a detailed description of the dataset details \& calibrations done for the Hoi! gripper depicted in Supplementary ~\cref{fig:gravitycomp}.

\subsection{Recording Device Specs}
We show the sensor specifications of the recording devices in ~\cref{tab:sensor_specs}.
\begin{table}[h]

\centering
\setlength{\tabcolsep}{2pt}
\resizebox{0.8\linewidth}{!}{%
\begin{tabular}{l|ccc}
\hline
\diagbox{Plat.}{Specs}
& \textbf{Cam/FOV} 
& \textbf{Res.} 
& \textbf{Rate} \\ \hline

Aria 
& 2 / 70--150$^\circ$ 
& 1408$^2$
& 30Hz \\

ZED 
& 2 / 90$^\circ$ 
&  1280x720 
& 20Hz \\

Digit 
& 1 / -- 
& VGA 
& 20Hz \\

F/T 
& -- 
& 6-axis 
& 100 Hz \\

iPhone 
& 1 / 77$^\circ$ 
& 4K/LiDAR 
& 30Hz \\

Spot 
& 5 / 360$^\circ$ 
& Gray/RGB,D 
& 15Hz\\

GoPro 
& 1 / 90-122$^\circ$ 
& 4K  
& 60Hz \\

\hline
\end{tabular}}
\caption{Key sensor specifications in the dataset.}
\label{tab:sensor_specs}

\end{table}

Regarding the accuracy of the force-torque measurements we refer to the datasheet. Key specs include an accuracy across all axes of $< 2\%$ and a noise-free resolution @ 100 Hz of 70-100 mN / 0.6-2.1 mNm.

\begin{figure}[h]
    \centering
\includegraphics[width=\linewidth]{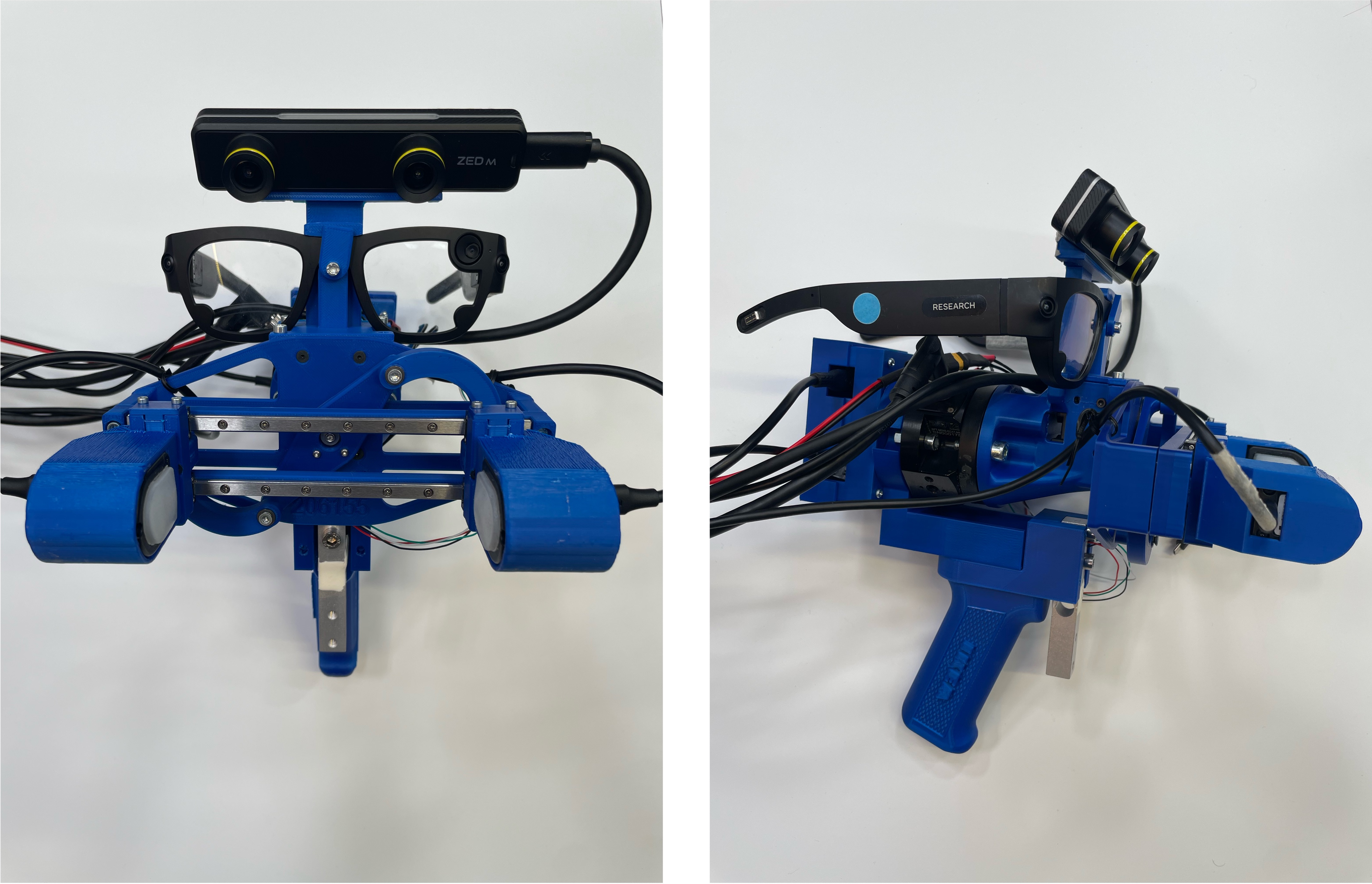}
    \caption{\textbf{Hoi! Gripper.} The 2-finger parallel gripper is operated through the load cell, where the measured load is translated into gripping force. Interaction force and tactile contact pressure are measured through the Digit and Force-Torque sensors respectively. Aria Glasses and a stereo camera provide pose estimation and wrist-view observations. We will release the design as open source.}
    \label{fig:gravitycomp}
    \smallskip
\end{figure}

\subsection{Motor Calibration}
The Hoi! gripper's gripping force is modeled as 
\[F^{(\mathrm{grip})}_{i} = g(J(q),\,\eta(I),\,I)\]
where $I$ denotes the motor current, $J(q)$ the gripper Jacobian as a function of the motor position $q$, and $\eta(I)$ a load-dependent efficiency factor. 
First, the motor torque is expressed as a proportional function of the current

\[
\tau(I_{\mathrm{mA}}) 
= k_1 I + k_2 .
\]
with $k_1 = 1.769$ and $k_2 = -0.2214$, as stated in the data sheet. Second, the gripping force is derived via the jacobian of the kinematic relationship $F = J(q)\,\tau$, where $\tau$ is calculated as a function of the lever angle $q$:

\[
J(q)
=
2\!\left(
- L_1 \sin q
-\frac{L_1^{\,2} \sin q \cos q}
       {\sqrt{\,L_2^{\,2} - \big(L_1 \sin q\big)^{2}\,}}
\right).
\]

To account for efficiency variations due to load-dependent motor performance and friction, we introduce a load-dependent calibration factor. 
This factor is obtained empirically by gripping a force sensor and recording pairs of measured gripping forces and corresponding motor currents. Using least-squares estimation, we determine $\eta(I)$ across multiple load regimes.

\subsection{Inter-Sensor Calibration}
We use Kalibr~\cite{rehder2016kalibr} to calibrate the gripper in the following order: We first calibrate the intrinisics of the ZED camera, and then the visual-inertial extrinsics to the Zed's IMU. We then use visual-inertial calibration between the ZED images and the IMU in the force-torque sensor to find the extrinsics between stereo camera and FT sensor. The Aria device runs its own intrinsic calibration, and we find the extrinsics through stereo calibration between one Aria and one Zed camera in their overlapping field-of-view

\begin{figure*}[t!]
    \centering
\includegraphics[width=\linewidth]{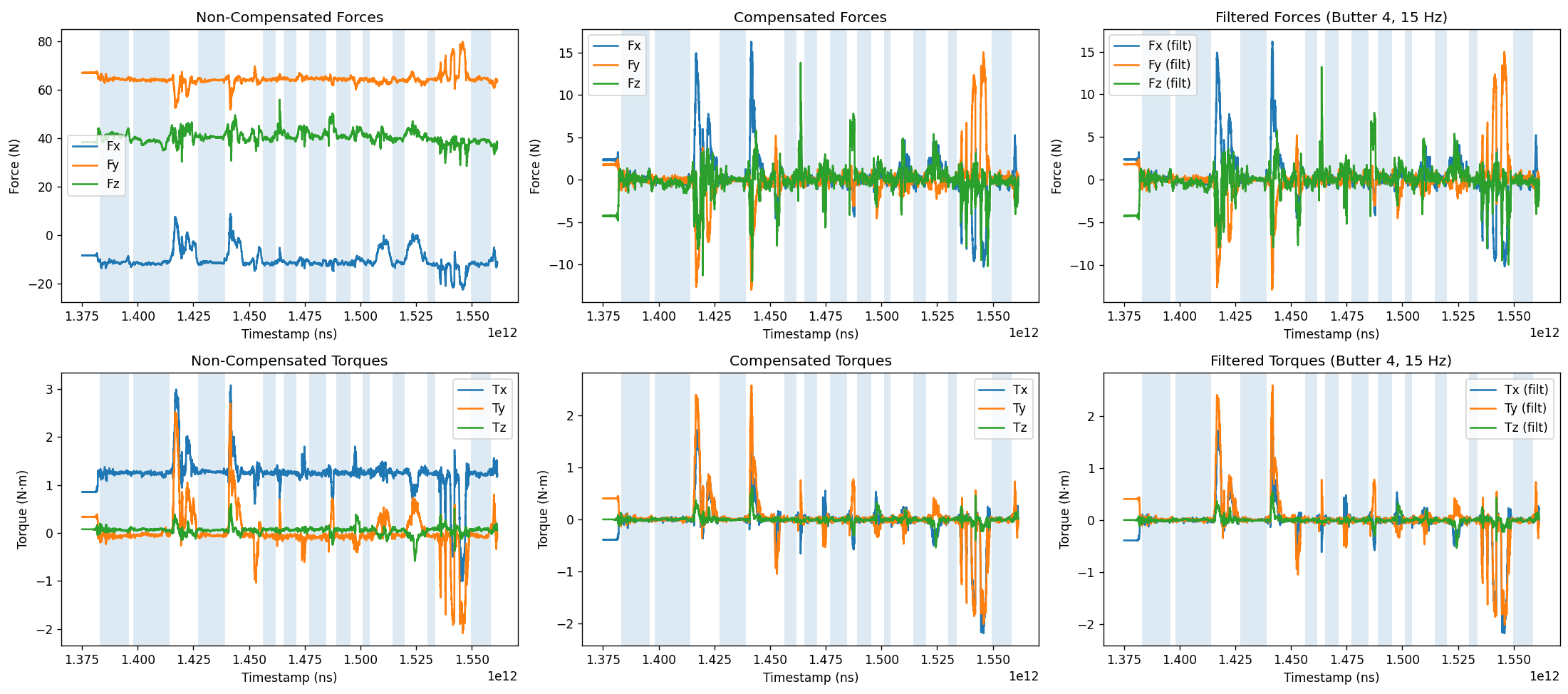}
    \caption{\textbf{Gravity Compensation.} We depict the uncompensated (left), compensated (middle) and filtered compensated (right) forces (upper row) and torques (lower row) of an examplary gripper recording (bathroom\_2). The no-load windows are stylized in blue.}
    \label{fig:gripper}
    \smallskip
\end{figure*}

\subsection{Gripper Gravity Compensation}
To measure the isolated interaction forces between gripper and furniture, we need to compensate for gravitational forces acting on the endeffector, as well as internal biases of the FT sensor. 
The governing equation is:
\[
\begin{aligned}
\mathbf{f}_S^{\text{meas}} &= \mathbf{f}_S^{\text{ext}} + \mathbf{f}_S^{g} + \mathbf{b}_f, \\
\boldsymbol{\tau}_S^{\text{meas}} &= \boldsymbol{\tau}_S^{\text{ext}} + \boldsymbol{\tau}_S^{g} + \mathbf{b}_\tau,
\end{aligned}
\]
where
\[
\begin{aligned}
\mathbf{f}_S^{g} &= \mathbf{R}_{S\leftarrow W}\, m\, \mathbf{g}_W, \\[3pt]
\boldsymbol{\tau}_S^{g} &= \mathbf{r}_{C\leftarrow S} \times \mathbf{f}_S^{g}, \\[3pt]
\boldsymbol{\tau}_S^{\text{ext}} &= \mathbf{r}_{P\leftarrow S} \times \mathbf{f}_S^{\text{ext}},
\end{aligned}
\]

where $\mathbf{r}_{S\!P}$ is the vector from the sensor origin to the contact point.
Here, $\mathbf{f}_S^{\text{meas}}$ and $\boldsymbol{\tau}_S^{\text{meas}}$ are the measured forces and torques in the sensor frame $S$, $\mathbf{f}_S^{\text{ext}}$ and $\boldsymbol{\tau}_S^{\text{ext}}$ are the external forces and torques acting on the sensor, $\mathbf{f}_S^{g}$ and $\boldsymbol{\tau}_S^{g}$ are the gravitational forces and torques acting on the sensor, $\mathbf{b}_f$ and $\mathbf{b}_\tau$ are the internal biases of the force-torque sensor, $\mathbf{R}_{S\leftarrow W}$ is the rotation matrix from the world frame $W$ to the sensor frame $S$, $m$ is the mass of the endeffector assembly, $\mathbf{g}_W$ is the gravity vector in the world frame, and $\mathbf{r}_{C\leftarrow S}$ is the vector from the sensor origin to the center of mass of the endeffector assembly.
We measure the mass of the endeffector using a scale, while the center of mass is estimated from the CAD model of the endeffector. $\mathbf{R}_{S\leftarrow W}$ is taken from the Aria SLAM and extrinsic calibration.
Internal biases are estimated during no-load conditions, where external forces and torques are zero. We estimate no load measurement windows by subtracting the median filtered force magnitude from the raw force magnitude and thresholding the result. We then estimate the biases by solving the following least-squares problem:

\[
\min_{\mathbf{b}_f, \mathbf{b}_\tau} \sum_{k=1}^{N} \left\| 
\begin{bmatrix}
\mathbf{f}_{S,k}^{\text{meas}} - \mathbf{R}_{WS,k}^\top\, m\, \mathbf{g}_W - \mathbf{b}_f \\
\boldsymbol{\tau}_{S,k}^{\text{meas}} - \mathbf{r}_{S\!C} \times (\mathbf{R}_{WS,k}^\top\, m\, \mathbf{g}_W) - \mathbf{b}_\tau
\end{bmatrix} \right\|^2,
\]
where $N$ is the number of no-load samples.
Given the estimated biases and the known mass and center of mass, we can now compute the external forces and torques during interaction as
\[
\begin{aligned}
\mathbf{f}_S^{\text{ext}} &= \mathbf{f}_S^{\text{meas}} - \mathbf{R}_{WS}^\top\, m\, \mathbf{g}_W - \mathbf{b}_f, \\
\boldsymbol{\tau}_S^{\text{ext}} &= \boldsymbol{\tau}_S^{\text  {meas}} - \mathbf{r}_{S\!C} \times (\mathbf{R}_{WS}^\top\, m\, \mathbf{g}_W) - \mathbf{b}_\tau.
\end{aligned}
\].
The results of gravity compensation are depicted in Supplementary ~\cref{fig:gravitycomp}. We also apply a Butterworth filter of degree 4 to filter noise.

\section{Alignment of Sensors in the Hoi! Dataset Recordings}
\begin{figure*}[t]
    \centering
\includegraphics[width=\linewidth]{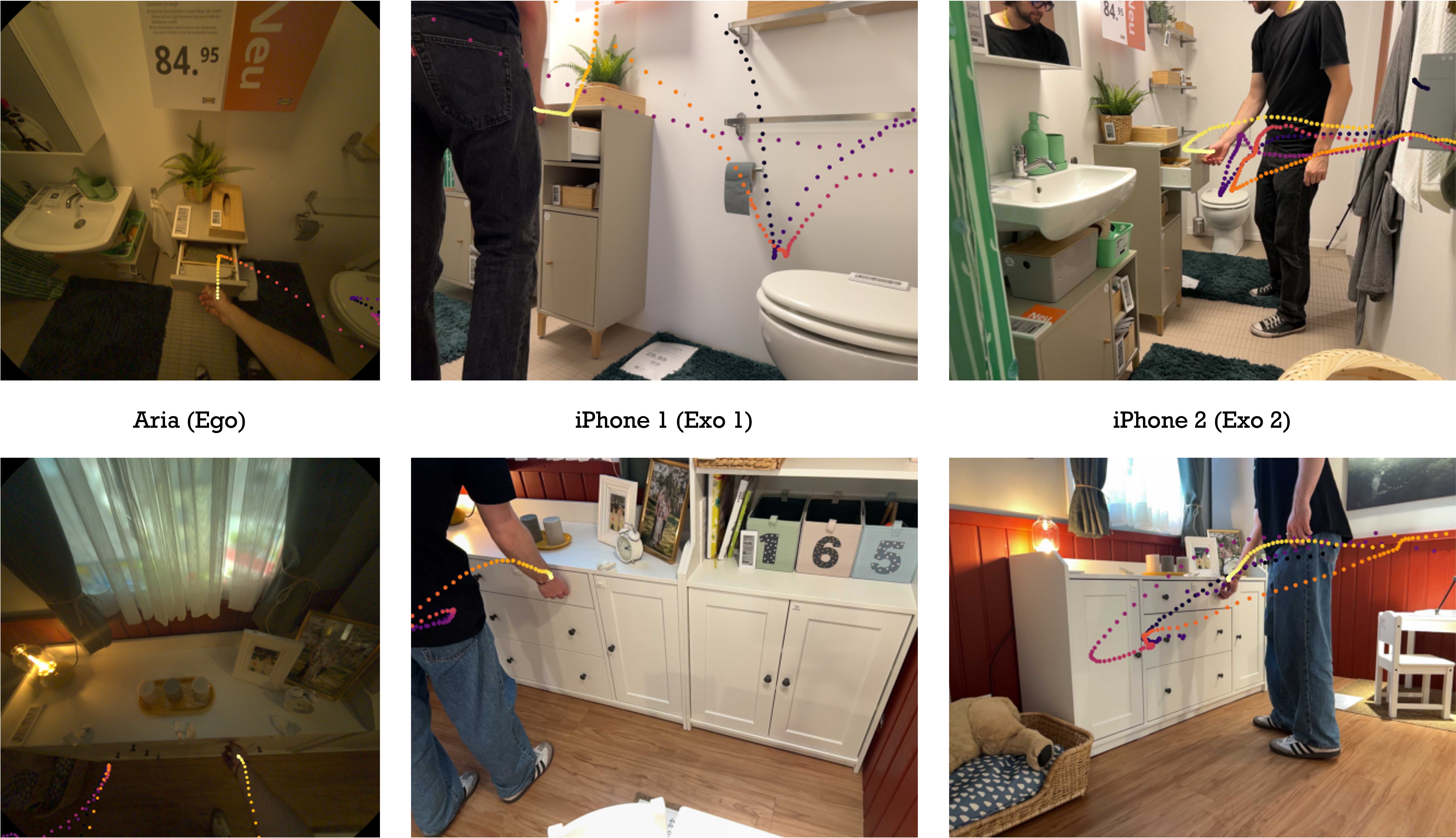}
    \caption{\textbf{Spatial and Temporal Alignment} We both temporally and spatially align all recording modules. This allows us to capture interactions over time across multiple viewpoints.}
    \label{fig:egoexo}
    \smallskip
\end{figure*}

We give a more in-depth explantion of both temporal and spatial alignment of the multiple sensor streams in the Hoi! dataset. This process allows us to capture interactions over time and across multiple perspectives, as depicted in
~\cref{fig:egoexo}.

\subsection{Time Alignment}

\begin{figure}[t]
    \centering
\includegraphics[width=\linewidth]{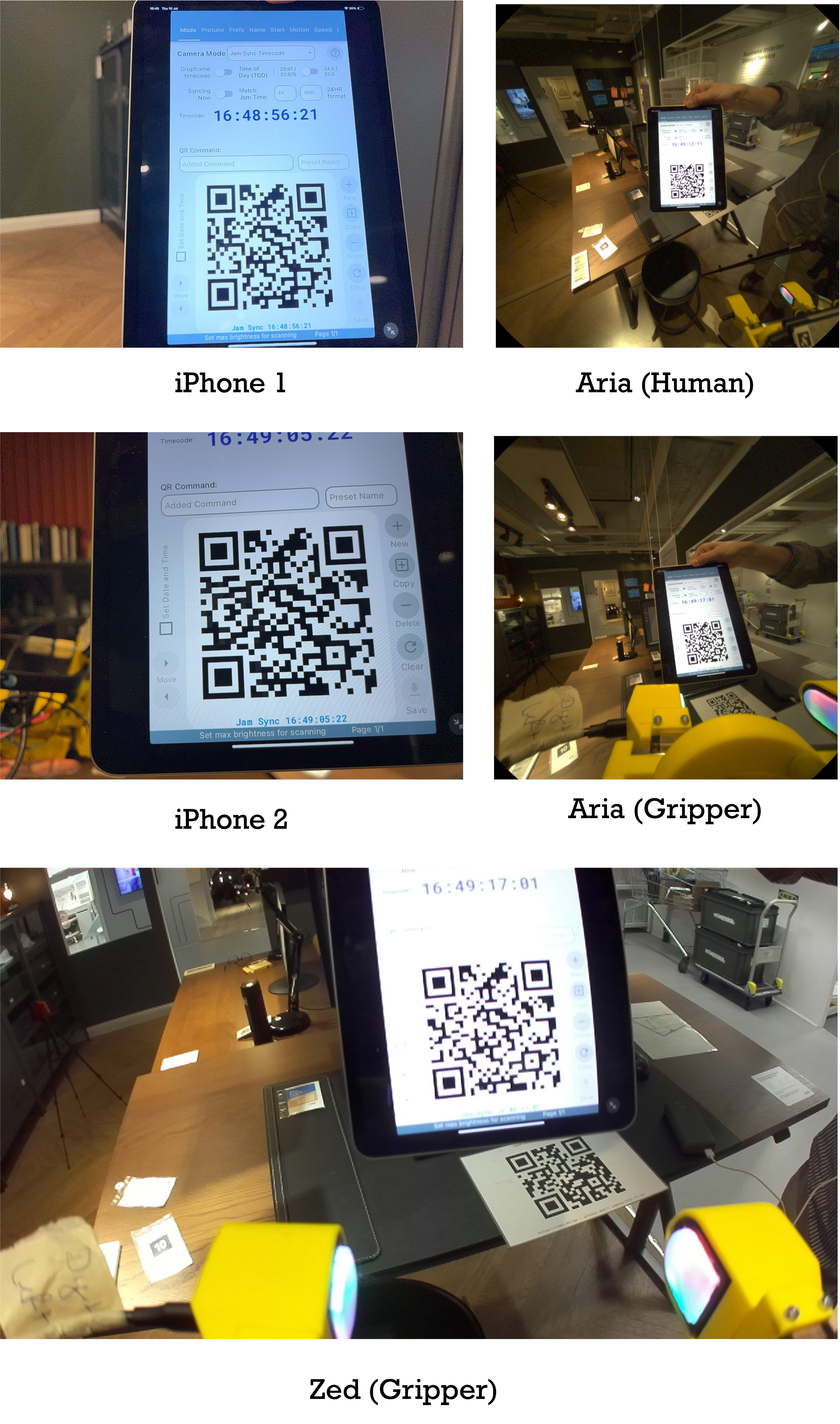}
    \caption{\textbf{Time Alignment.} The recording is started for all recording modules of a single recording and a QR code encoding the current time is shown to all video streams, so that the individual clocks can be aligned in post processing.}
    \label{fig:timealign}
    \smallskip
\end{figure}
As the different recording modules run independently of each other on different internal clocks, we need to align the time frames of the recordings in post processing. During recording, we display a QR code encoding the current Unix timestamp at 25 Hz into each camera stream $i$. 
We detect and decode the first QR code, yielding a time pair $(t_{internal}, t_{\text{ref}})$, where $t_{internal}$ is the internal timestamp of the video stream $i$ at which the QR code was captured, and $t_{\text{ref}}$ is the corresponding reference Unix timestamp encoded in the QR code.
Detecting and decoding this QR code yields a time offset for each video stream with respect to a common reference time, in our case the Aria egocentric camera stream, as \[
\Delta_i = (t_{\text{ref},i} - t_{\text{internal},i}) - (t_{\text{ref,Aria}} - t_{\text{internal,Aria}}),
\]
where $\Delta_i$ denotes the relative clock offset of stream~$i$ with respect to the Aria reference. The uncertainty of this single-shot estimate is dominated by frame-timing quantization: assuming the QR update occurs uniformly $\delta_i \sim \mathrm{Uniform}(0, T_i), i \in \{\text{ArEgo},\, \text{Sensor}\}$,
within the exposure of the first detected frame, the standard deviation of the alignment error is
\[
\sigma_{\Delta_i} = \sqrt{\tfrac{T_i^2}{12} + \tfrac{T_{\text{Aria}}^2}{12}},
\]
yielding a $95\%$ confidence interval of approximately $\pm 1.96\,\sigma_{\Delta_i}$.
For typical frame rates of $30{-}60\,\text{Hz}$, this corresponds to a temporal alignment accuracy of roughly $10{-}25\,\text{ms}$ per stream.
In a representative example, when the iPhone records at $60~\text{Hz}$ ($T_i \approx 16.7~\text{ms}$) and the Aria at $30~\text{Hz}$ ($T_{\text{Aria}} \approx 33.3~\text{ms}$), the resulting uncertainty is
$\sigma_{\Delta_i} \approx 10.8~\text{ms}$,

\subsection{Spatial Alignment}
We spatially align all recording devices into a common reference frame using visual localization against the high-resolution 3D scans. We first construct a reference set of dense 2D-3D correspondences from the point cloud and corresponding panoramic images captured with the scanner. 
We rectify the panoramic images into multiple perspective images with virtual camera parameters $\mathbf{K}, \mathbf{R}, \mathbf{t}$. We further convert the point cloud into a mesh using the Leica proprietary software and render depth maps from the same virtual camera poses. The rendered depth maps are then back-projected into 3D space using the known virtual camera intrinsics~$\mathbf{K}$ and extrinsics~$(\mathbf{R}, \mathbf{t})$ as
\[
\mathbf{P}
= \mathbf{R}^{-1} \mathbf{K}^{-1}
\begin{bmatrix}
u\,d \\[2pt]
v\,d \\[2pt]
d
\end{bmatrix}
- \mathbf{R}^{-1}\mathbf{t},
\]
where $(u,v)$ are pixel coordinates in the image plane and $d$ is the corresponding depth value at that pixel. This operation transforms each depth pixel into a 3D point~$\mathbf{P}$ in the global coordinate frame, thereby establishing dense 2D-3D correspondences between the rendered depth maps and the rectified panoramic images.
We assume zero drift and global consistency of the per-device SLAM, which reduces the global registration problem to a single rigid 3D-3D alignment. While in principle a single localized frame would suffice for this alignment, we automatically select a set of high-quality keyframes for robust registration. 
To obtain these keyframes, we filter all frames of a trajectory according to feature density, sharpness, and scene depth. Specifically, we retain frames with a high number of ORB~\cite{rublee2011orb} keypoints, a high variance of the Laplacian (indicating low motion blur) and a high mean estimated depth. With the latter, we filter out frames where the device is very close to the furniture, where usually no good references are in view. For Aria and GoPro cameras, which do not provide depth maps, we use DepthAnything~v2~\cite{yang2024depthv2}. On this filtered subset, we extract DINOv2~\cite{oquab2024dinov2} features and perform farthest-point sampling~\cite{gonzalez1985clustering,qi2017pointnetdeephierarchicalfeature} in feature space to select $N$ diverse and representative keyframes per trajectory.
We estimate the 6-DoF pose of each keyframe
through hloc~\cite{sarlin2019coarse} with the 2D-3D database constructed from the Leica scan and robustly estimate a single rigid transformation $\mathbf{T}_{\text{world}}^{\text{query}}$ between each sensor trajectory and the shared world frame.
Since we also have the corresponding query poses $\mathbf{T}_{\text{query}}^{\text{cam}_i}$, with $\text{query} \in \{\text{iPhone}, \text{Aria}, \text{UMI}\}$, we can estimate a single rigid transformation $\mathbf{T}_{\text{world}}^{\text{query}}$ aligning each query trajectory to the common world frame by
\[
 \mathbf{T}_{\text{world}}^{\text{query}} = \mathbf{T}_{\text{world}}^{\text{cam}_i} \, {\mathbf{T}_{\text{query}}^{\text{cam}_i}}^{-1}.
  \]
After localizing each keyframe, we compute the Frobenius distances between all pairwise combinations of the $N$ estimated transformations $\mathbf{T}_{\text{world}}^{\text{query}}$ and reject outliers based on a threshold relative to the median distance. 
We then average the remaining inlier transformations to obtain a final robust estimate of $\mathbf{T}_{\text{world}}^{\text{device}}$. 
This simple outlier rejection strategy is sufficient, as Hierarchical Localization already performs RANSAC~\cite{fischler1981ransac}-based PnP pose estimation internally.
Finally, we transform all Aria poses and hand poses into the common world frame using the estimated transformation $\mathbf{T}_{\text{world}}^{\text{Aria}}$, yielding globally aligned 6-DoF trajectories for the Aria head, hands, and Hej gripper. The iPhone cameras are statically mounted, and we therefore directly measure their poses in the world frame during spatial calibration. The UMI gripper poses are transformed into the world frame using the estimated transformation $\mathbf{T}_{\text{world}}^{\text{UMI}}$.

\section{Spot Recordings}
\begin{figure}[t]
    \centering
\includegraphics[width=\linewidth]{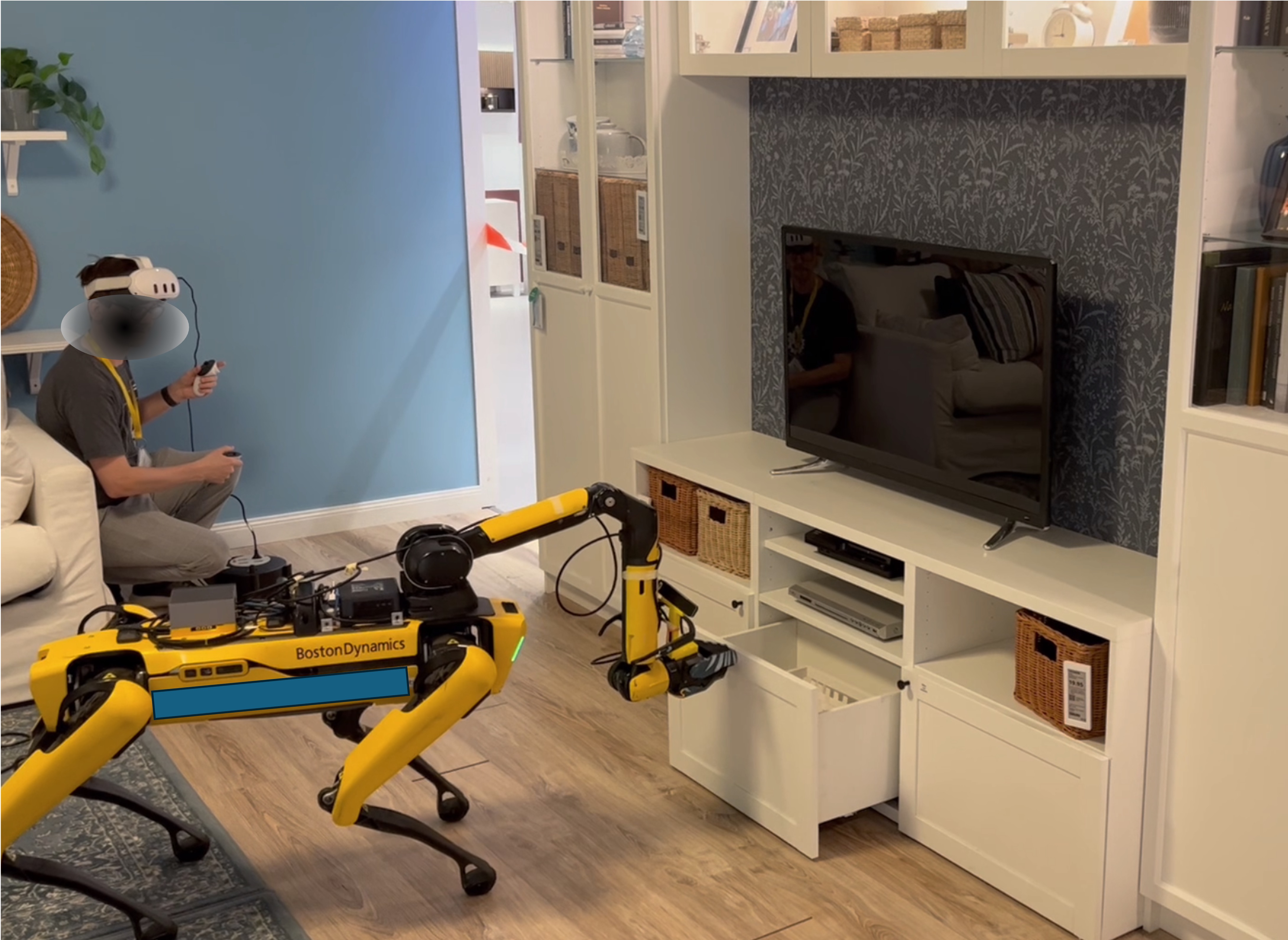}
    \caption{\textbf{Spot Teleoperation} The Spot Robot is teleoperated using a Meta Quest 3. We retarget the remote's 6-DoF pose to the gripper and control the base using the remote mounted Joystick.}
    \label{fig:spot_teleop}
    \smallskip
\end{figure}

As mentioned in the main paper, we collect robotic data  for a subset of interactions using a Boston Dynamics Spot Robot. Here, we record the robot's joint states, surrounding RGB-D cameras, the gripper RGB-D camera as well as Aria data using a wrist-mounted Aria, imitating the wrist viewpoint of gripper and wrist recordings. As shown in ~\cref{fig:spot_teleop}, the robot is teleoperated using a Meta Quest 3. Here the operator controls the robot base using the joystick on the Quest remote, while the remote's 6-DoF pose is retargeted to the Spot gripper. The opening angle is also controlled via button on the remote.  

\section{Evaluations}
In the following we give more in-depth description of how we created the evaluation groundtruth used in our evaluations from the Hoi! dataset.
\subsection{In-The-Wild Articulation Estimation}
In this section, we evaluate a set of different approaches to articulation estimation. This task concerns recovering articulation parameters (axis and position) as well as the articulation type from visual observations. We first evaluate ArtiPoint~\cite{werby2025articulatedobjectestimationwild}, a recent method for articulation estimation in the wild. We also include the Gaussian-Splatting-based approach ArtGS \cite{liu2025artgsbuildinginteractablereplicas}. Furthermore, we investigate how GPT-5, as a state-of-the-art vision-language model (VLM), can infer articulation types from both egocentric and exocentric observations. For this analysis, we make use of our dataset's posed RGB frames (both egocentric and exocentric), articulation annotations, and provided 3D ground-truth.As the Aria does not provide dense depth, we generate dense RGB-D data, we employ MapAnything~\cite{keetha2025mapanythinguniversalfeedforwardmetric} to predict dense depth for each posed frame in our interaction sequences. We convert these depth estimates into metric scale by rendering depth maps from our 3D ground-truth meshes under the corresponding camera pose. We then robustly compute a global scale factor by forming a per-pixel scale histogram and selecting the scale as the mode
\[s = \arg\max_{s} \, h(s)\], 
where $h(s)$ denotes the histogram of per-pixel ratios between predicted and rendered depths. This is necessary because certain regions in the predicted depth differ from the rendered depth (\eg, the operator's hand or articulated parts present in the predicted depth but absent in the mesh rendering). We finally apply the scale factor to obtain dense, metrically accurate depth for each frame. The actual ground truth articulation parameters are provided using a light-weight manual annotation tool presented in \cite{werby2025articulatedobjectestimationwild}. 

\subsection{Tactile Force Estimation}
We evaluate the utility of our dataset for the task tactile force estimation from gel-based tactile sensors. This task aims to estimate contact forces acting on the sensor's surface solely from the tactile images captured by the sensor during contact. We specifically focus on estimating the normal and tangential forces, as these are most relevant for manipulation tasks. We combine the tafigctile images provided by the Hoi! gripper's Gelsight Digit sensors with the corresponding forces provided by the gripper's force-torque sensor and the gripper's gripping forces into ground-truth labels. We evaluate two versions the Sparsh model~\cite{higuera2024sparsh}, the SOTA method for self-supervised tactile representations. We evaluate both the DINO~\cite{caron2021dino} and DINOv2~\cite{oquab2024dinov2} decoder, with a fine-tuned force estimation decoder (referred to as 'Task 1' in the original Sparsh paper).
The interaction forces during grasping can be decomposed into two components: a normal preload $F^{(\mathrm{grip})}_{i}$ resulting from the motor-torque–induced gripping force, and an external reactive force $F^{(\mathrm{ext})}_{i}$ exerted by the environment. The external force is measured by the force–torque (FT) sensor as $F^{(\mathrm{ext})}_{s}$ in the sensor’s local coordinate frame. The preload $F^{(\mathrm{grip})}_{i}$, while sensed by the tactile sensors (Digits), represents an internal force and is therefore not captured by the FT sensor.
To express all forces in a consistent coordinate system, we define an interaction frame $i$, whose axes are aligned with the Digit frames. The interaction frame is defined to be coaxial with the Digit frames; however, since the two Digits face each other, corresponding axes have opposing directions ($x_{\mathrm{L}} = -x_{\mathrm{R}}$, $z_{\mathrm{L}} = -z_{\mathrm{R}}$, $y_{\mathrm{L}} = y_{\mathrm{R}}$). This does not affect our analysis, as we only consider the sum of absolute force magnitudes. The FT-sensor measurements are rotated into this frame as
$F^{(\mathrm{ext})}_{i} = \mathbf{R}_{i \leftarrow s} \, F^{(\mathrm{ext})}_{s}$,
where $\mathbf{R}_{i \leftarrow s}$ denotes the rotation from the sensor frame $s$ to the interaction frame $i$. All subsequent equations and force components are defined in this frame.
Considering the hardware configuration of the \textit{Hoi} gripper, the combined absolute forces are computed as
\begin{equation}
\begin{aligned}
  F^{(\mathrm{tang})}_{i}
    &= \Bigl\| \sum_{k \in \{\mathrm{L},\,\mathrm{R}\}} |F^{(\mathrm{ext})}_{i,\,k,\{x,y\}}| \Bigr\|_2, \\[3pt]
  F^{(\mathrm{norm})}_{i}
    &= \sum_{k \in \{\mathrm{L},\,\mathrm{R}\}} |F^{(\mathrm{ext})}_{i,\,k,z}|, \\[3pt]
  F^{(\mathrm{comb})}_{i}
    &= \sqrt{\bigl(F^{(\mathrm{tang})}_{i}\bigr)^2 + \bigl(F^{(\mathrm{norm})}_{i}\bigr)^2}.
\end{aligned}
\label{eq:force-components}
\end{equation}

The gripping force is estimated from the gripper’s torque–current relationship, its Jacobian, and a load-dependent calibration factor as $F^{(\mathrm{grip})}_{i} = g(J(q),\,\eta(I),\,I)$. Since the Sparsh models are trained on force data within the range of $[4,4,5]\,\mathrm{N}$ along the $x$, $y$, and $z$ axes, respectively, we clip the combined ground-truth magnitudes to $F^{(\mathrm{norm})}_{\max} = 10\,\mathrm{N}$ and $F^{(\mathrm{tang})}_{\max} = \sqrt{32}\,\mathrm{N}$ to ensure consistency between training and evaluation distributions. This avoids extrapolation to unseen force magnitudes and provides a fair comparison of model performance.
The gripping force is estimated using the gripper's torque-current relationship, its Jacobian and a load dependent calibration factor $F_{\mathrm{grip}} = g\!\left(J(q),\, \eta(I),\, I\right)$. 
As the Sparsh models are trained on force data within therange of $[4, 4, 5]\,\mathrm{N}$ for $x$, $y$, and $z$, respectively, we clip our combined ground-truth magnitudes to $F_{\mathrm{max,normal}} = 2 \times 5 = 10\,\mathrm{N}$ and $F_{\mathrm{max,tangential}} = \sqrt{4^2 + 4^2} = \sqrt{32}\,\mathrm{N}$ to ensure consistency between training and testing distributions. This avoids extrapolation to unseen force magnitudes and provides a fair assessment of model performance. The extended evaluation results are depicted in ~\cref{tab:tactile_forces}.
\begin{table*}[]
    \centering
    \caption{\textbf{Tactile Force Estimation.} We show the evaluation results for our tactile force estimation evaluation per location and split into tangential, normal and combined forces over 2 digit images.}
    \resizebox{1.0\linewidth}{!}{%
  \begin{tabular}{l c c c c c c}
    \toprule
    & \multicolumn{2}{c}{Tangential} & \multicolumn{2}{c}{Normal} & \multicolumn{2}{c}{Combined} \\
    \cmidrule(lr){2-3}\cmidrule(lr){4-5}\cmidrule(lr){6-7}
    \textbf{Location} & DINO & DINOv2 & DINO & DINOv2 & DINO & DINOv2 \\
    \midrule
    bathroom\_2   & 2.07 [1.88, 2.25] & 2.02 [1.81, 2.23] & 4.81 [4.54, 5.09] & 4.97 [4.69, 5.26] & 4.92 [4.64, 5.21] & 5.17 [4.88, 5.47] \\
    bedroom\_4    & 3.23 [3.09, 3.37] & 2.77 [2.63, 2.92] & 3.22 [3.07, 3.38] & 3.72 [3.60, 3.85] & 3.41 [3.22, 3.60] & 3.96 [3.81, 4.11] \\
    bedroom\_6    & 3.39 [3.19, 3.59] & 3.77 [3.52, 4.00] & 3.48 [3.23, 3.71] & 3.43 [3.22, 3.63] & 4.31 [4.04, 4.57] & 4.45 [4.24, 4.66] \\
    kitchen\_7    & 3.05 [2.82, 3.27] & 3.83 [3.67, 4.00] & 2.76 [2.60, 2.92] & 3.50 [3.39, 3.61] & 3.46 [3.28, 3.65] & 3.19 [2.97, 3.41] \\
    office\_1     & 3.63 [3.41, 3.86] & 3.76 [3.52, 4.01] & 3.88 [3.62, 4.14] & 4.07 [3.86, 4.27] & 4.61 [4.30, 4.92] & 4.91 [4.67, 5.17] \\
    livingroom\_1 & 2.54 [2.31, 2.77] & 2.61 [2.40, 2.82] & 3.63 [3.34, 3.91] & 3.71 [3.46, 3.95] & 3.89 [3.56, 4.21] & 3.90 [3.61, 4.19] \\
    \midrule
    \textbf{Overall} 
      & \textbf{3.07} [2.87, 3.26] & \textbf{3.18} [2.99, 3.38] &
        \textbf{3.45} [3.24, 3.66] & \textbf{3.79} [3.61, 3.96] &
        \textbf{3.86} [3.62, 4.11] & \textbf{4.11} [3.90, 4.33] \\
    \bottomrule
  \end{tabular}}
    \label{tab:tactile_forces}
\end{table*}

\begin{figure*}[t!]
    \centering
\includegraphics[width=\linewidth]{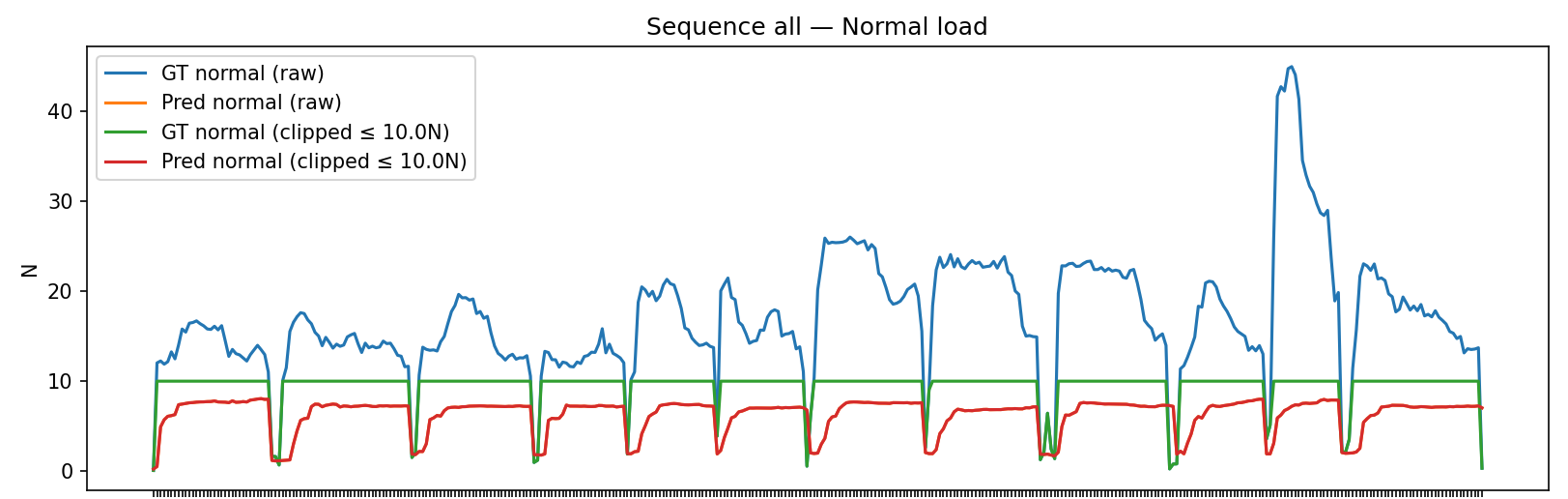}
    \caption{\textbf{Tactile force Estimation.} We depict the GT forces, clipped GT forces and the estiamted forces (DINOv2) for an examplary recording.}
    \label{fig:tactileforceestimation}
    \smallskip
\end{figure*}

As depicted in ~\cref{fig:tactileforceestimation}, we observe that the force predictions generally under-predict the tactile forces, even though the GT is clipped to be within the training range. 

\subsection{Visual Force Estimation}
We evaluate the utility of our dataset to the task of visual force estimation. In this task a 3D interaction force is estimated alongside an affordance (interaction type) in order to complete a given manipulation goal. In our context the input might be an RGB-D image of a drawer alongside the prompt "open the drawer" and the model would predict where to interact and what force to apply. We evaluate the ForceSight model \cite{collins2023forcesighttextguidedmobilemanipulation}, a model that aims to predict forces as part of visual-force goals for robotic manipulation, demonstrating that force goals can significantly increase robotic manipulation performance.
The ForceSight dataset consists of interaction sequences that include posed RGB-D observations, per-frame force–torque (FT) readings, and gripping forces. Each sequence is paired with an open-language goal derived from our interaction annotations. Using the Hoi! gripper’s FT sensor, we generate per-image force–torque labels and leverage the 3D ground-truth trajectories provided by our dataset. We evaluate the model across a diverse subset of six environments. Because raw force-torque signals may include operator-induced forces in directions unrelated to the articulation (\eg, internal stresses not required for the intended motion), we report results on both the raw data and a motion-aligned variant. For fair comparison, we project the measured force vector $\mathbf{f}$ onto the gripper's linear velocity $\mathbf{v}$ using $f_{\parallel} = \mathbf{f} \cdot \frac{\mathbf{v}}{\lVert \mathbf{v} \rVert}$, and similarly project the torque vector $\boldsymbol{\tau}$ onto the rotational velocity $\boldsymbol{\omega}$ as $\tau_{\parallel} = \boldsymbol{\tau} \cdot \frac{\boldsymbol{\omega}}{\lVert \boldsymbol{\omega} \rVert}$. This removes force and torque components that do not contribute to the articulated interaction, resulting in a fairer evaluation signal.

\subsection{Multimodal Learning for Real-World Robotics}
We conduct a force-centric experiment to demonstrate \textbf{Hoi!}'s utility for multimodal skill learning. Specifically, we train a visual force predictor on \textbf{Hoi!} gripper data and integrate the learned force prior into Spot's position-force controller. Given an RGB-D observation and the articulated-part class label, the model predicts a feed-forward interaction force $\mathbf{f}_{ff}$.

Our predictor uses frozen DINOv2 image features and a small MLP head to estimate a residual on top of a class-specific force prior. More formally, given an observation $o=(I,d,c)$ with RGB image $I$, depth-derived features $d$, and class label $c$, the model predicts
\begin{equation}
\hat{\mu} = \mu_c + \alpha \, h_\theta(\phi_{\text{DINO}}(I)),
\end{equation}
where $\mu_c$ denotes the class prior in log-force space, $\phi_{\text{DINO}}$ is the frozen visual backbone, $h_\theta$ is the learned regression head, and $\alpha$ scales the visual residual. We train the network to regress the log-transformed peak interaction force using an $\ell_2$ loss,
\begin{equation}
\mathcal{L} = \left\| \hat{\mu} - \log(1+F) \right\|_2^2 .
\end{equation}
At test time, the predicted force prior is mapped back to force space and injected as a feed-forward term into the controller.

As a baseline, we use standard impedance control,
\begin{equation}
\mathbf{f}_{cmd} = \mathbf{K}\,\Delta \mathbf{r} + \mathbf{D}\,\Delta \mathbf{v},
\end{equation}
where $\Delta \mathbf{r}$ and $\Delta \mathbf{v}$ denote pose and velocity errors, respectively. In the position-force variant, we augment this controller with the predicted feed-forward force term:
\begin{equation}
\mathbf{f}_{cmd} = \mathbf{K}\,\Delta \mathbf{r} + \mathbf{D}\,\Delta \mathbf{v} + \mathbf{f}_{ff}.
\end{equation}

We evaluate on articulated objects in our lab with increasing mechanical stiffness. For each trial, the robot is initialized in front of the object, and a simple geometric heuristic selects the articulation type (eccentric handle $\rightarrow$ revolute, centric handle $\rightarrow$ prismatic), such that failures are predominantly attributable to interaction force selection rather than articulation misclassification. Each articulated part is operated five times. As shown below, the learned force prior substantially improves real-world success rate (SR), highlighting \textbf{Hoi!}'s potential for multimodal policy learning in real-world manipulation settings.

\begin{table}[h]
\centering
\setlength{\tabcolsep}{4pt}
\renewcommand{\arraystretch}{1.1}
\resizebox{0.8\linewidth}{!}{%
\begin{tabular}{lccc}
\hline
\textbf{Control Type} & \textbf{Drawer} & \textbf{Oven} & \textbf{Dishwasher} \\
\hline
Position (SR [$\%$])                & 100 & 0  & 0  \\
Position-Force (SR [$\%$])       & 100 & 80 & 60 \\
\hline
Avg. Force Pred. [N]    & 16,80  & 46,38 & 53,29 \\
\hline
\end{tabular}}
\label{tab:control_sr_force}
\vspace{-14pt}
\end{table}

\subsection{Hand Pose Estimation}
\begin{table}[h!]
   \centering
   \footnotesize
   \caption{\textbf{Hand Pose Estimation} Average PCK@0.15 on our evaluation locations and compared to the Hamer baseline performace on New Days, VISOR and Ego4D datasets.}
   \label{tab:pck-ego-only}
   \setlength{\tabcolsep}{5pt}
   \begin{tabular}{l c}
     \toprule
     \textbf{Location / Dataset} & \textbf{PCK} \\
     \midrule
     bathroom\_2    & 0.757 \\
     bedroom\_4     & 0.764 \\
     bedroom\_6     & 0.708 \\
     kitchen\_7     & 0.535 \\
     office\_1      & 0.732 \\
     livingroom\_1  & 0.748 \\
     \midrule
     \textbf{Overall} & \textbf{0.699} \\
     \midrule
     \textbf{New Days \cite{newdays}} & 0.888 \\
     \textbf{VISOR \cite{darkhalil2022epickitchensvisorbenchmarkvideo}}        & 0.893 \\
     \textbf{Ego4D \cite{grauman2022ego4d}}        & 0.844 \\
     \bottomrule
   \end{tabular}
\end{table}
As our dataset captures not only gripper interactions but also hand interactions, we additionally explore the performance of hand-pose estimation across these viewpoints. Because the Aria MPS hand keypoints are automatically generated-derived from stereo and globally optimized trajectories but not from manual annotations or motion-capture systems, we treat this evaluation as an exploratory analysis rather than a definitive benchmark.

We employ the method of Pavlakos et al.~\cite{pavlakos2024reconstructing}, which has shown strong in-the-wild performance. Using the Aria MPS trajectories, we project hand keypoints into egocentric frames and compute the commonly used PCK metric \cite{pck}.

While the results (~\cref{tab:pck-ego-only}) show a noticeable performance gap relative to controlled benchmarks, this is expected given the challenging characteristics of our real-world setting - fast hand motions, natural manipulation behaviors, and lower-resolution egocentric imagery. Rather than indicating deficiencies, these findings highlight the difficulty of egocentric manipulation scenes and underline the opportunity for future methods to better leverage the rich multimodal signals present in our dataset.


\end{document}